\pdfoutput=1

\documentclass[11pt]{article}

\usepackage[final]{acl}

\usepackage{times}
\usepackage{latexsym}
\usepackage{amsmath}
\usepackage{amssymb}
\usepackage{mathtools}

\usepackage{hyperref}
\usepackage{url}
\usepackage{booktabs}
\usepackage{wrapfig}
\usepackage{colortbl}  
\usepackage{xcolor}  
\usepackage{multirow}
\usepackage{caption}
\usepackage{listings}
\usepackage{rotating}
\usepackage{bm}
\usepackage{enumitem}
\usepackage{etoc}
\usepackage{tabularx}
\usepackage{titlesec}
\usepackage{pifont}
\usepackage{makecell}

\usepackage[T1]{fontenc}

\usepackage[utf8]{inputenc}

\usepackage{microtype}

\usepackage{inconsolata}

\usepackage{graphicx}

\newcommand{\cmark}{\ding{51}}%
\newcommand{\xmark}{\color{lightgray} \ding{55}}%

\title{Outlier-Safe Pre-Training for Robust 4-Bit Quantization \\ of Large Language Models}

\author{Jungwoo Park$^{1,2}$, Taewhoo Lee$^{1,2}$, Chanwoong Yoon$^1$ \\ \textbf{Hyeon Hwang$^1$, Jaewoo Kang$^{1,2}$\thanks{Corresponding author}} \\
  $^1$Korea University, $^2$AIGEN Sciences \\
  \{jungwoo-park, taewhoo, cwyoon99, hyeon-hwang, kangj\}@korea.ac.kr
}

\begin{document}
\maketitle
\begin{abstract}

Extreme activation outliers in Large Language Models (LLMs) critically degrade quantization performance, hindering efficient on-device deployment. While channel-wise operations and adaptive gradient scaling are recognized causes, practical mitigation remains challenging. We introduce \textbf{Outlier-Safe Pre-Training (OSP)}, a practical guideline that proactively prevents outlier formation rather than relying on post-hoc mitigation. OSP combines three key innovations: (1) the Muon optimizer, eliminating privileged bases while maintaining training efficiency; (2) Single-Scale RMSNorm, preventing channel-wise amplification; and (3) a learnable embedding projection, redistributing activation magnitudes originating from embedding matrices. We validate OSP by training a 1.4B-parameter model on 1 trillion tokens, which is the first production-scale LLM trained without such outliers. Under aggressive 4-bit quantization, our OSP model achieves a 35.7 average score across 10 benchmarks (compared to 26.5 for an Adam-trained model), with only a 2\% training overhead. Remarkably, OSP models exhibit near-zero excess kurtosis (0.04) compared to extreme values (1818.56) in standard models, fundamentally altering LLM quantization behavior. Our work demonstrates that outliers are not inherent to LLMs but are consequences of training strategies, paving the way for more efficient LLM deployment.
The source code and pretrained checkpoints are available at
\url{https://github.com/dmis-lab/Outlier-Safe-Pre-Training}.
\end{abstract}

\section{Introduction}

\begin{figure}[t]
\centering
\includegraphics[width=1.0\linewidth]{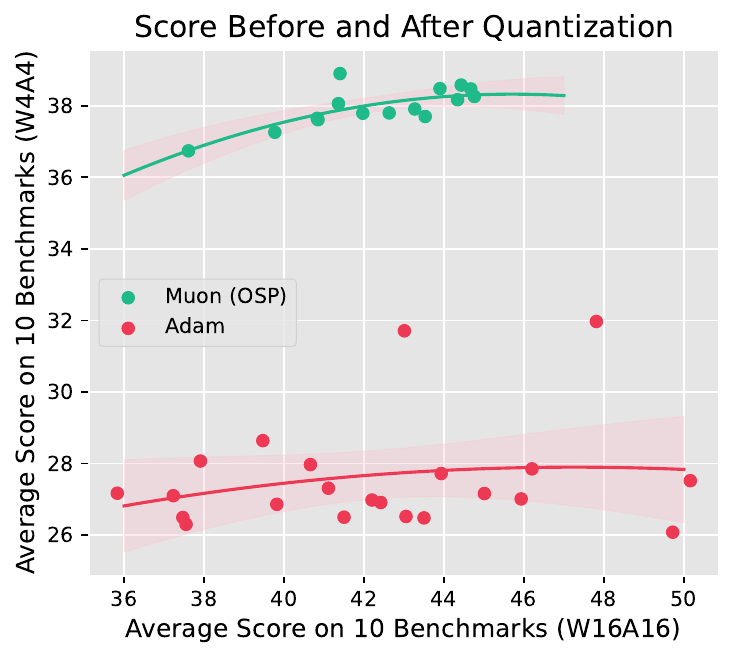}
\caption{
Comparison of 
performance degradation patterns under 4-bit
quantization across different training methodologies. \textbf{Adam} represents various open-source LLMs trained with the Adam optimizer. Muon (OSP) represents checkpoints from our model trained within the \textbf{Outlier-Safe Pre-Training (OSP)} framework.
Results demonstrate that our framework produces fundamentally different degradation characteristics
compared to conventional Adam-trained models.
}
\vspace*{-1em}
\label{fig:quantization-law}
\end{figure}
\begin{figure*}[t]
\centering

\vspace{-0.8em}
\begin{minipage}[b]{0.3\linewidth}
    \begin{flushleft}
        \includegraphics[width=1.0\linewidth]{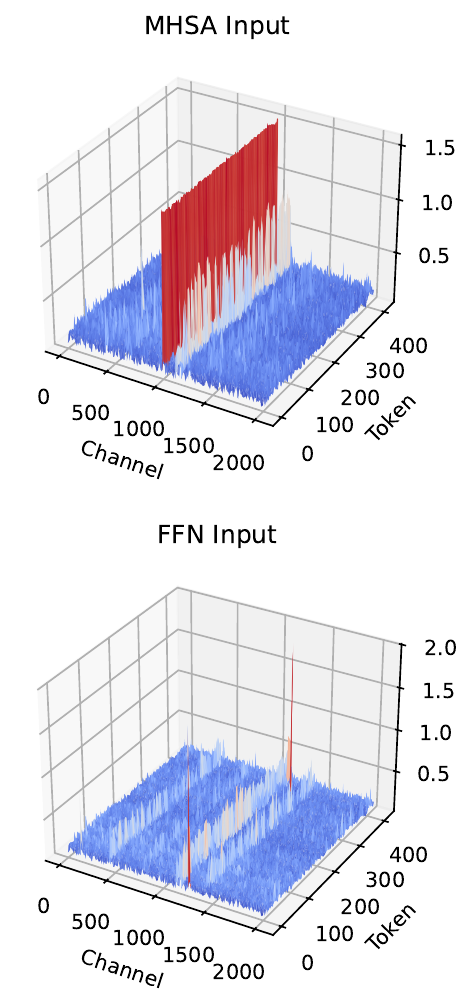}
    \end{flushleft}
  \centerline{\textbf{(a) Adam}}
\end{minipage}
\hspace{1em}
\begin{minipage}[b]{0.3\linewidth}
    \begin{flushleft}
        \includegraphics[width=1.0\linewidth]{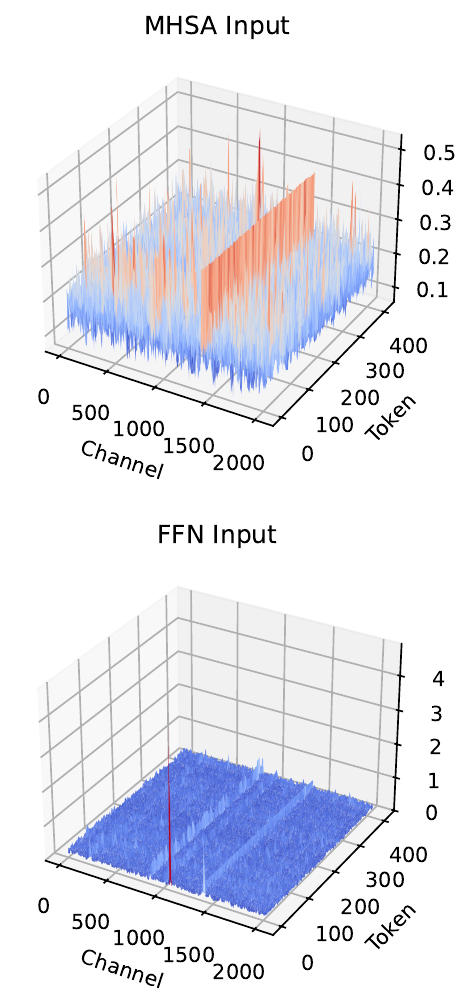}
    \end{flushleft}
  \centerline{\textbf{(b) Muon}}
\end{minipage}
\hspace{1em}
\begin{minipage}[b]{0.3\linewidth}
    \begin{flushleft}
        \includegraphics[width=1.0\linewidth]{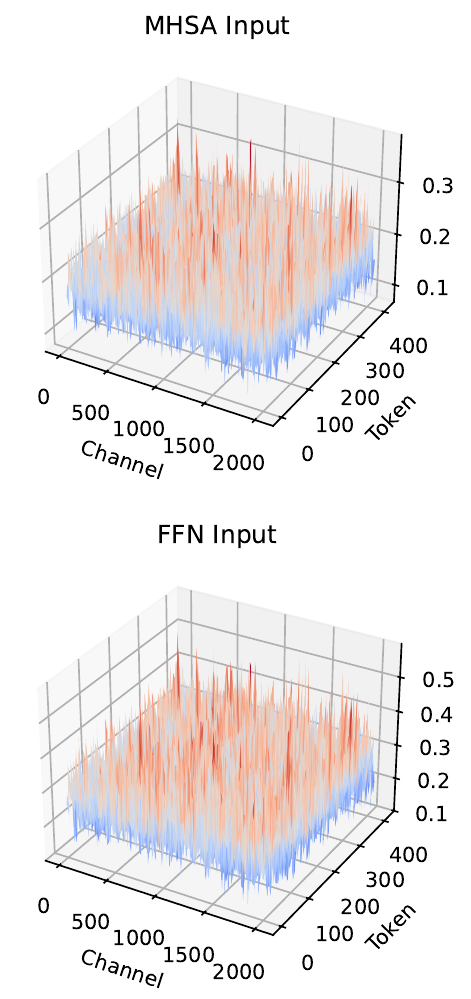}
    \end{flushleft}
  \centerline{\textbf{(c) Muon (OSP)}}
\end{minipage}
\caption{
Activation distribution analysis from the 20th layer input to Multi-Head Self-Attention (MHSA) and Feed-Forward Network (FFN) modules in 1.4 billion parameter models trained on 100 billion tokens. Three optimization strategies are examined: (a) standard Adam optimizer, (b) Muon optimizer without architectural modifications, and (c) the \textbf{Outlier-Safe Pre-Training (OSP)} framework. The histograms reveal that substituting Adam with Muon optimizer alone provides insufficient mitigation of activation outliers.
}
\label{fig:act-hist-100B-comp}
\end{figure*}

Quantization has emerged as a practical solution for deploying Large Language Models (LLMs) in resource-constrained environments \citep{dettmers2022llmint8}. As modern LLMs often scale to hundreds of billions of parameters, reducing the bit-width of weights and activations can significantly decrease memory consumption during inference. However, the pervasive presence of outliers within LLM architectures presents a fundamental obstacle to effective low-bit quantization.

Contemporary LLMs trained with standard optimization techniques invariably develop outlier features during pre-training, posing persistent challenges for quantization \citep{bondarenko2023quantizabletransformers, he2024OP}. Post-Training Quantization (PTQ) methods have gained substantial attention for mitigating these outliers at inference time, enabling immediate deployment without requiring costly retraining.
While PTQ offers a pragmatic solution,
it remains a reactive rather than preventive approach, implicitly accepting outliers as an intrinsic property of LLMs instead of addressing their root cause.

This raises a critical question: are outliers an inevitable consequence of large-scale language model training? Returning to first principles, the ideal solution for robust quantization would be to train LLMs in a way that inherently prevent outlier formation during pre-training. Several recent studies have investigated the origins of outlier features in LLMs, suggesting three distinct perspectives on their root causes: channel-wise scaling factors in normalization layers \citep{kovaleva-etal-2021-bert-busters, wei2022outliersuppression, he2024OP}, the attention sink phenomena \citep{bondarenko2023quantizabletransformers, guo2024activedormantattnheads, gu2025attnsinkemerges}, and diagonal optimizers such as Adam and AdaFactor \citep{he2024OP, guo2024activedormantattnheads}. While each study has proposed methods to mitigate outliers at the pre-training stage based on these findings, existing approaches remain preliminary, lacking consideration for production-level scalability and practicality.
Most are limited to models with fewer than one billion parameters or training corpora of under 100 billion tokens, and often overlook the computational overhead of alternative optimization strategies as well as compatibility issues introduced by architectural modifications.

Therefore, we present the \textbf{Outlier-Safe Pre-Training (OSP)} framework, a practical guideline that integrates existing findings to enable the development of quantization-friendly models at industrial scale. Our approach prioritizes three key objectives: (1) scaling to production-level training requirements, (2) maintaining computational efficiency comparable to standard methods, and (3) ensuring complete architectural compatibility with existing inference pipelines.

Our empirical results demonstrate that the model trained with the \textbf{OSP} framework remains free of outlier features over the course of training on one trillion tokens.
Our methodology reduces memory usage by 33\% compared to standard approaches and maintains competitive performance, with only a 2\% increase in training time.
Under aggressive 4-bit weight and activation quantization, our model significantly outperforms comparable open-source alternatives. Moreover, our approach is orthogonal to existing PTQ techniques, enabling complementary performance gains when combined with post-training quantization.

Beyond quantization performance, our work provides new insights into outlier-free model behavior. Notably, our analysis reveals that attention sinks persist even in the absence of outliers, suggesting they are not inherently responsible for outlier formation.
While \citet{bondarenko2023quantizabletransformers} has attributed outliers to attention sinks implemented via extreme negative attention logits, our outlier-free models exhibit similar behavior via concentrated positive attention on specific tokens, without the emergence of massive activations.
This finding challenges existing assumptions about the relationship between attention mechanisms and outliers. By publicly releasing the first production-scale outlier-free LLM, we enable further investigation into how the absence of outliers affects other emergent properties of large language models.

\section{Related Works}

\subsection{Quantization}
\label{sec:related-works-quantization}

Quantization reduces the precision of weights and activations by mapping continuous floating-point values to discrete integers. For an $n$-bit quantization, floating-point value $x$ is mapped to a discrete integer representation through:  
\begin{equation}
    \hat{x}_\text{int}=\text{clip}\left(\Big\lfloor \frac{x}{s} \Big\rceil + z, 0, 2^n-1\right),
\end{equation}
where $s$ denotes the scale factor and $z$ represents the zero-point offset. The presence of outliers catastrophically inflates the scale factor, resulting in severe rounding errors and substantial information loss in low-bit settings.

To mitigate this challenge, the research community has pursued two primary directions. Quantization-Aware Training (QAT) enhances model robustness to quantization during the training process \citep{liu2024llmqat, efficientqat}.
However, rather than eliminating outliers directly, QAT simulates quantization errors during training to improve tolerance to discretization errors.
Moreover, this simulation significantly impedes training, making it impractical for large-scale deployment.

In contrast, Post-Training Quantization (PTQ) improves quantization performance of pre-trained models without additional training. PTQ encompasses various techniques: allocating higher precision exclusively to outlier channels \citep{dettmers2022llmint8, kim2024squeezellm}, leveraging the Hessian of quantization error for optimal weight rounding \citep{frantar-gptq}, and utilizing activation statistics \citep{xiao2023smoothquant, lin2024awq}.
A particularly notable advancement involves applying random rotation matrices, which effectively redistribute outliers across different channels for both activations and weights without requiring architectural modifications \citep{chee2023quip, tseng2024quipsharp, ashkboos2024quarot}.

Despite their sophistication, these PTQ methods fundamentally accept outliers as an intrinsic characteristic of LLMs, operating under the assumption that specific outlier channels persist across all token positions. Rather than preventing outlier formation, PTQ techniques represent sophisticated workarounds that manage the symptoms while leaving the underlying cause unaddressed.

\subsection{Massive Activations and Attention Sinks}
\label{sec:related-works-attention-sinks}

Meanwhile, recent studies have identified a distinct class of outliers in LLMs, referred to as \textit{massive activations}~\citep{sun2024massiveactivations}. Unlike conventional outliers that appear in specific channels, massive activations exhibit sporadic behavior, emerging unpredictably in particular layers and predominantly within start tokens or delimiter tokens~\citep{sun2024massiveactivations, barbero2025whyllmsattendfirsttoken}.

Emerging evidence indicates a strong correlation between massive activations and \textit{attention sinks}~\citep{xiao2024efficientstreaming}, where attention layers excessively focus on specific tokens throughout the sequence. \citet{bondarenko2023quantizabletransformers} posit that attention layers occasionally perform partial residual state updates by assigning attention weights to semantically unimportant tokens, effectively implementing a "no-op" operation. To achieve near-zero softmax values for most tokens, models drive attention logits toward negative infinity, inadvertently generating massive activations as a byproduct of this cancellation.

This understanding has motivated several research directions aimed at mitigating attention sinks to address quantization challenges. These approaches primarily target the attention mechanism through architectural modifications \citep{bondarenko2023quantizabletransformers, guo2024activedormantattnheads, gu2025attnsinkemerges}
or adjustments to key-value caches \citep{liu-etal-2024-intactkv, son-etal-2024-prefixing-attn-sinks}. Nevertheless, empirical results demonstrate only limited performance gain,
and growing evidence suggests that massive activations and outlier features underlie different mechanisms \citep{sun2024massiveactivations, guo2024activedormantattnheads}.


\subsection{Privileged Bases and Diagonal Optimizers}
\label{sec:related-works-optimizers}

An alternative interpretation of outlier emergence in neural networks centers on the concept of \textit{privileged bases}~\citep{elhage2023privilegedbases}. 
Element-wise operations such as non-linear activation functions and channel-wise scaling factors in normalization layers induce a form of basis alignment wherein certain channels disproportionately accumulate magnitude. This observation has motivated previous efforts to manipulate
normalization layers for outlier mitigation \citep{kovaleva-etal-2021-bert-busters, wei2022outliersuppression, he2024OP}.

Remarkably, recent studies suggest that the primary cause of outlier formation may not lie in architectural design, but rather in the adaptive gradient scaling mechanisms of
optimizers \citep{elhage2023privilegedbases, caples2024adamprivileged, he2024OP, guo2024activedormantattnheads}. Optimizers such as \mbox{RMSProp}, \mbox{Adam}, and \mbox{AdaFactor} maintain running statistics of per-parameter gradient variance, applying element-wise standardization during parameter updates. This scaling strategy,
often called diagonal preconditioning,
introduces a preferential basis that can encourage outlier emergence.

Second-order optimizers that leverage Hessian matrix
are proposed as alternatives to diagonal optimizers.
By incorporating loss landscape curvature rather than
rescaling gradient elements,
these methods achieve faster convergence than first-order optimizers. K-FAC~\citep{martens2015kfac} approximates the Hessian matrix using the Fisher information and constructs block-diagonal approximations layer-wise through Kronecker factorization instead of computing curvature for all parameters. Shampoo~\citep{gupta2018shampoo} and SOAP~\citep{vyas2024soap} further reduce computational complexity by introducing separate preconditioners for each tensor dimension.


\section{Outlier-Safe Pre-Training}

While recent investigations have proposed explanations for the emergence of outliers in LLMs, translating these insights into production-scale solutions remains challenging. Architectural modifications that eliminate privileged bases, including gating modules~\citep{bondarenko2023quantizabletransformers} or the normalization layer removal~\citep{he2024OP}, break compatibility with established inference frameworks.

Similarly, non-diagonal optimizers still impose prohibitive computational costs even with various approximation techniques. Our experiments show that second-order methods require three times the memory footprint of Adam and reduce training throughput by 25\% (see Table~\ref{tab:optimizer-speed}). Such overheads make them impractical for trillion-token scale pre-training scenarios.

We present the \textbf{Outlier-Safe Pre-Training (OSP)} framework, which synthesizes prior insights while operating within production constraints. Our approach comprises three components that maintain the training costs and preserve the canonical transformer design, ensuring compatibility with existing inference systems and post-training techniques for complementary performance gains.

\subsection{Integration of the Muon Optimizer}

\begin{table}
\centering




\resizebox{1.0\columnwidth}{!}{

\begin{tabular}{lcccc}

\toprule
\textbf{Optimizer} & \textbf{TPS (Relative)} & \textbf{Memory Usage} & \textbf{Build Time} \\
\midrule

Adam & 4.07M (100\%) & $O\left(36LD^2\right)$ & 2m 30s \\
Muon & 3.99M (97.9\%) & $O\left(24LD^2\right)$ & 3m 48s \\
$\text{Shampoo}^\dagger$ & 3.07M (75.5\%) &  $O\left(\displaystyle\frac{338}{3}LD^2\right)$ & 24m 24s \\
SOAP & -- & $O\left(\displaystyle\frac{302}{3}LD^2\right)$ & $\geq$ 3 hours \\

\bottomrule

\end{tabular}

}

\caption{
Training throughput comparison across different optimizers estimated on TPU-v4 512 Pod Slice infrastructure. \textbf{TPS} denotes tokens processed per second during training operations. Theoretical memory usage requirements and code compilation time (\mbox{\textbf{Build Time}}) on JAX~\citep{jax2018github} framework are reported.
}
\vspace*{-1em}
\label{tab:optimizer-speed}
\end{table}

The foundation of our framework rests upon the integration of the Muon optimizer~\citep{jordan2024muon}, which represents a fundamental departure from diagonal preconditioning paradigms. Using the Newton-Schulz algorithm~\citep{schulz1933iterativenewtonschulz, higham2008newtonschulz}, Muon iteratively transforms the gradient matrix to approximate orthogonalization according to:
\begin{equation}
    G=U\Sigma V^T \mapsto UV^T,
\end{equation}
where $U \in \mathbb R^{m \times m}$ and $V \in \mathbb{R}^{n \times n}$ are the singular vectors, and $\Sigma \in \mathbb{R}^{m \times n}$ is the diagonal matrix of singular values.

This algorithm distinguishes Muon from adaptive optimizers including Adam and AdaFactor. Muon maintains only gradient momentum without element-wise scaling and applies parameter updates through full-rank linear transformations. This algorithm eliminates privileged bases inherent to diagonal preconditioning, preventing systematic channel amplification and outlier occurrence.

Although technically classified as first-order optimization, Muon exhibits convergence properties comparable to second-order optimizers \citep{jordan2024muon}. Recent theoretical analyses \citep{bernstein2024oldoptnewnorm, bernstein2024modularduality, duvvuri2024caspr} have established that Muon's update rule converges to that of Shampoo in the limiting case where preconditioner accumulation is disabled.
This equivalence elucidates Muon's efficiency gains, which are achieved without explicit Hessian computation or gradient preconditioning, reaching 97.9\% of the training throughput of standard Adam (see Table~\ref{tab:optimizer-speed}).

Prior to our investigation, the scalability of Muon to billion-parameter architectures trained on trillion-token corpora has remained unexplored. We provide the first empirical validation demonstrating that Muon successfully scales to production-level training while maintaining its outlier-prevention properties throughout the pre-training process.

\subsection{Single-Scale RMSNorm}

Despite eliminating optimizer-induced privileged bases via the Muon optimizer, channel-wise scaling factors within normalization layers constitute an explicit basis alignment \citep{wei2022outliersuppression, he2024OP, nrusimha2024mitigating}, necessitating careful architectural modifications to achieve comprehensive outlier prevention.

\citet{qin2023transnormerllm} proposed Simple RMSNorm (SRMSNorm), which removes the entire learnable parameters and instead divides vectors by $\sqrt{d}$, where $d$ denotes the vector dimensionality. While this approach demonstrates effectiveness in preventing outlier emergence \citep{he2024OP}, our initial experiment reveals preliminary limitations that impede practical deployment. Specifically, dividing normalized vectors by $\sqrt{d}$ causes severe activation magnitude suppression during initial training phases, substantially slowing convergence. Conversely, fixing the scale to 1.0 leads to training instability as the model matures.

This phenomenon indicates that transformer architectures require dynamic scaling of activation magnitudes during training. To address these limitations, we propose \textbf{Single-Scale RMSNorm (\mbox{\textsc{SSNorm}})}, which introduces
to uniformly control activation magnitude across all dimensions. The \textsc{SSNorm} layer is defined as:
\begin{equation}
    {\textsc{SSNorm}}(x) = \gamma \, \frac{x}{\|x\|_2},
\end{equation}
where $\gamma \in \mathbb{R}$ represents the scaling parameter. 
This design enables adaptive adjustment of activation scales while eliminating the channel-wise multiplication. By constraining all dimensions to share a single scaling factor, \mbox{\textsc{SSNorm}} fundamentally prevents the emergence of privileged coordinates while maintaining stable optimization dynamics.

\subsection{Decoupled Embedding Optimization}
\label{sec:method-decoupled-emb-opt}
\begin{table*}[t]
\centering
\resizebox{1.0\textwidth}{!}{%
\begin{tabular}{l|cc|c|c|cc|cc|cc|cc|cc}
%

\toprule
  \multirow{2}{*}{\textbf{Optimizer}}
& \multirow{2}{*}{\textbf{\textsc{SSNorm}}}
& \multirow{2}{*}{\textbf{\textsc{EmbProj}}}
& \multirow{2}{*}{\textbf{Ex. Kurt.}}
& \multirow{2}{*}{\textbf{Had.}}
& \multicolumn{2}{c|}{\textbf{16-16-16}}
& \multicolumn{2}{c|}{\textbf{4-8-16}}
& \multicolumn{2}{c|}{\textbf{4-8-8}}
& \multicolumn{2}{c|}{\textbf{4-4-16}}
& \multicolumn{2}{c}{\textbf{4-4-4}}
\\
& & & &
& Avg. & PPL
& Avg. & PPL
& Avg. & PPL
& Avg. & PPL
& Avg. & PPL
\\
\midrule
\multirow{2}{*}{Adam} & \multirow{2}{*}{\xmark} & \multirow{2}{*}{\xmark} & \multirow{2}{*}{1818.56} & \xmark
& 41.5 & 11.4
& 39.7 & 21.6
& 39.7 & 21.6
& 26.5 & 1e5
& 26.8 & 8e4 \\

& & & & \cmark
& 41.5 & 11.4
& 40.2 & 22.3
& 40.3 & 22.3
& 27.2 & 3e4
& 26.9 & 3e4 \\

\midrule
\multirow{2}{*}{\makecell[l]{$\text{Muon}^\dagger$ \\ (w/o Adam)}} & \multirow{2}{*}{\xmark} & \multirow{2}{*}{\xmark} & \multirow{2}{*}{361.35} & \xmark
& 41.0 & 11.7
& 38.4 & 14.8
& 38.3 & 14.8
& 26.3 & 1e6
& 26.3 & 8e5 \\

& & & & \cmark
& 41.0 & 11.7
& 37.5 & 15.4
& 37.5 & 15.4
& 33.3 & 24.5
& 33.1 & 24.8 \\

\midrule
\multirow{2}{*}{Muon} & \multirow{2}{*}{\xmark} & \multirow{2}{*}{\xmark} & \multirow{2}{*}{1575.12} & \xmark
& 41.5 & 11.4
& 40.0 & 13.8
& 40.0 & 13.8
& 29.4 & 934.3
& 29.0 & 1e4 \\

& & & & \cmark
& 41.5 & 11.4
& 40.6 & 12.9
& 40.6 & 12.9
& 38.6 & 15.7
& 38.4 & 15.8 \\

\midrule
\multirow{2}{*}{Muon} & \multirow{2}{*}{\ding{51}} & \multirow{2}{*}{\xmark} & \multirow{2}{*}{66.69} & \xmark
& \textbf{41.8} & \textbf{11.2}
& \textbf{41.0} & 12.4
& \textbf{40.9} & 12.4
& 36.6 & 43.3
& 36.4 & 44.2 \\

& & & & \cmark
& \textbf{41.8} & \textbf{11.2}
& \textbf{40.8} & 12.2
& \textbf{40.8} & 12.2
& 38.6 & 33.7
& 38.3 & 34.1 \\

\midrule
\multirow{2}{*}{Muon} & \multirow{2}{*}{\xmark} & \multirow{2}{*}{\ding{51}} & \multirow{2}{*}{703.23} & \xmark
& 40.0 & 12.3
& 38.4 & 14.8
& 38.4 & 14.8
& 31.0 & 99.7
& 30.4 & 114.6 \\

& & & & \cmark
& 40.0 & 12.3
& 39.2 & 13.9
& 39.3 & 13.9
& 36.3 & 22.1
& 36.2 & 22.3 \\

\midrule
\multirow{2}{*}{\makecell[l]{Muon \\ (OSP)}} & \multirow{2}{*}{\ding{51}} & \multirow{2}{*}{\ding{51}} & \multirow{2}{*}{\textbf{0.04}} & \xmark
& 41.4 & \textbf{11.2}
& 40.6 & \textbf{12.2}
& 40.6 & \textbf{12.2}
& \textbf{37.9} & \textbf{19.4}
& \textbf{37.5} & \textbf{19.6} \\

& & & & \cmark
& 41.4 & \textbf{11.2}
& 40.5 & \textbf{12.1}
& 40.5 & \textbf{12.1}
& \textbf{39.1} & \textbf{13.4}
& \textbf{38.9} & \textbf{13.5} \\

\bottomrule

\end{tabular}
}
\caption{
Ablation study examining models trained on 100 billion tokens. \textbf{\textsc{SSNorm}} indicates single-scale RMSNorm integration, while \textbf{\textsc{EmbProj}} denotes models incorporating learnable embedding projection layers. \textbf{Ex. Kurt} represents excess kurtosis measurements, and \textbf{Had.} indicates online Hadamard transformation applied to Feed-Forward Network (FFN). Bit-width configurations (e.g., 16-16-16, 4-8-16) specify quantization precision for weights, activations, and key-value cache respectively. \textbf{Avg.} presents average performance across 10 LLM benchmarks, while \textbf{PPL} reports WikiText-2 perplexity. $^\dagger$Model configuration that disables decoupled embedding optimization by training with Muon optimizer without Adam optimization on embedding layers (Section~\ref{sec:method-decoupled-emb-opt}).
}
\vspace*{-0.5em}
\label{tab:ablation-100B-design-choices}
\end{table*}

The final component of our framework addresses the computational challenges posed by embedding layers in modern LLMs. As vocabulary sizes continue to expand, embedding matrices have grown to constitute a substantial fraction of model parameters. The high dimensionality of these matrices presents significant computational bottlenecks for
non-diagonal optimizers.
Our empirical analysis reveals that applying orthogonalization to embedding
layers incurs an additional 6\% throughput degradation beyond the baseline computational cost.

To address this computational challenge, our framework maintains Adam optimization exclusively for the embedding layers.
This approach aligns with \citet{jordan2024muon}, who demonstrate that decoupling embedding matrices from Muon and training them with Adam achieves better convergence properties. For all experiments, we adopt decoupled optimization as the default configuration for embedding layers within the \textsc{OSP} framework.

Since this decoupling strategy potentially reintroduces outliers through the embeddings,
our framework further introduces a learnable full-rank embedding projection, positioned after the embedding layer and before the unembedding layer, which we refer to as \textsc{EmbProj}.
These matrices, inspired by PTQ methods \citep{chee2023quip, ashkboos2024quarot, liu2024spinquant}, redistribute any emerging outliers across different dimensions, preventing their concentration and propagation through other layers. The matrices can be absorbed into their adjacent embeddings after training, maintaining computational invariance~\citep{ashkboos2024slicegpt}. We employ orthogonal initialization to maintain the initial norm distribution of embedding vectors, thereby preserving training dynamics from initialization.

\section{Experiments}

\subsection{Outlier Quantification}
Prior to conducting our experimental evaluation, we establish a systematic approach for quantifying outlier emergence within model activations. Following established practices in the previous literature~~\citep{he2024OP, caples2024adamprivileged}, we
employ \textit{excess kurtosis} to quantify the degree of outlier concentration:
\begin{equation}
\text{Kurt}[X] - 3 = \mathbb{E} \left[ \left( \frac{X - \mu}{\sigma} \right)^4 \right] - 3,
\end{equation}
where $\mu$ and $\sigma$ denote the mean and standard deviation of activation $X$, respectively. This metric effectively captures the heavy-tailed nature of distributions containing outliers, with higher values indicating greater outlier presence.

\subsection{Experimental Setup}
\label{sec:experimental-setup}

We evaluate our framework through comprehensive experiments on 1.4B-parameter LLaMA~\citep{touvron2023llama} architecture. Our experimental design examines quantization performance across different optimizers and architectural modifications to validate the effectiveness of our approach at scale.

We assess model performance using perplexity on WikiText-2~\citep{merity2016wikitext2} and accuracy across 10 downstream benchmarks: ARC~\citep{allenai:arc}, CommonsenseQA~\citep{talmor2019csqa}, GSM8k (8-shot)~\citep{cobbe2021gsm8k}, HellaSwag~\citep{zellers2019hellaswag}, MMLU~\citep{hendryckstest2021mmlu}, OpenBookQA~\citep{mihaylov2018openbookqa}, PIQA~\citep{Bisk2020piqa}, SIQA~\citep{sap2019siqa}, TriviaQA (5-shot)~\citep{joshi2017triviaqa}, and WinoGrande~\citep{Sakaguchi2020winogrande}. This extensive evaluation protocol ensures thorough assessment of both the quantization robustness and the general capabilities of models trained with the {OSP} framework.

\subsection{Ablation Study}
\begin{figure*}[t]
\centering
\hspace*{-1em}
\includegraphics[width=0.95\textwidth]{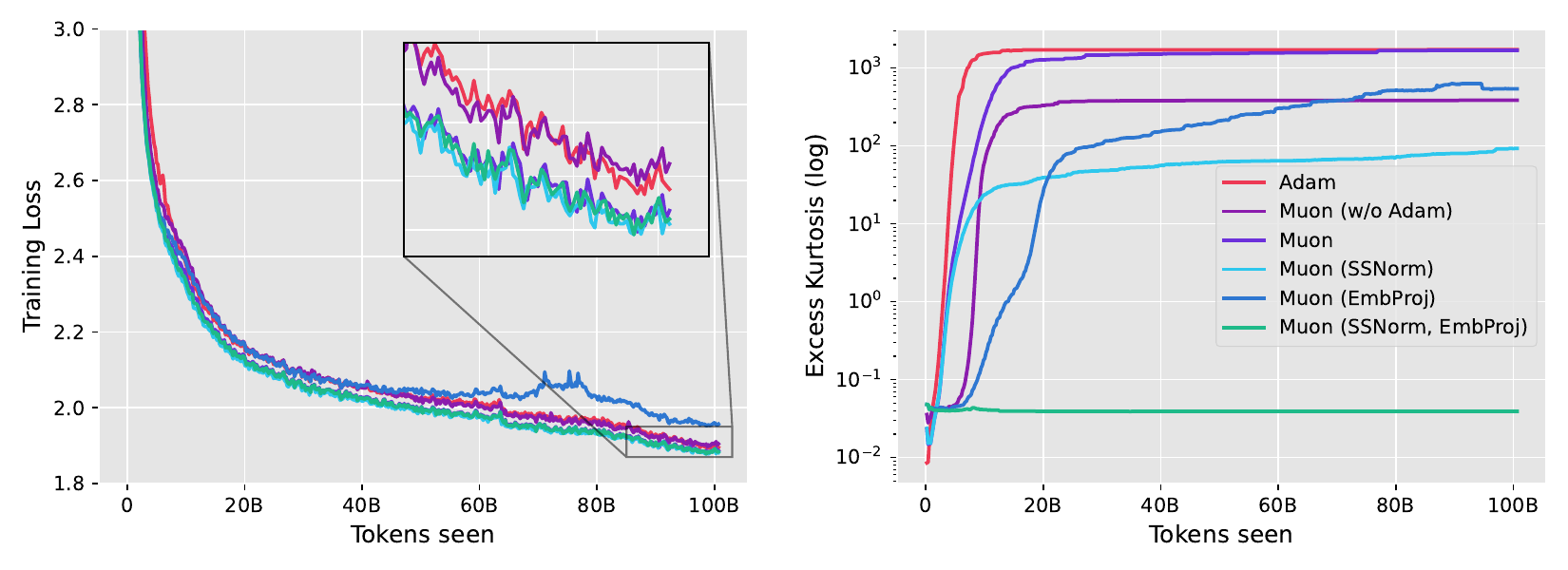}
\vspace*{-1em}
\caption{
Training dynamics comparison showing loss (left) and excess kurtosis evolution (right) across 100 billion tokens for ablation study examining various configurations of {OSP} components. The excess kurtosis demonstrate that only when all {OSP} components are simultaneously enabled does the kurtosis remain near zero throughout training, indicating complete elimination of outlier formation. Partial implementation of OSP components results in insufficient outlier suppression, as evidenced by elevated kurtosis values that persist across the training duration.
}
\vspace*{-0.5em}
\label{fig:100B-training-log}
\end{figure*}
\begin{figure}[t]
\centering
\hspace*{-1em}
\includegraphics[width=1\linewidth]{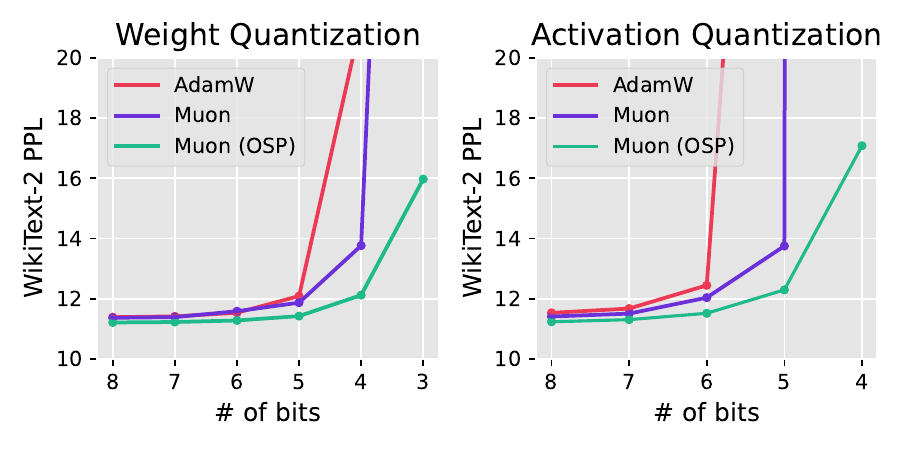}
\vspace*{-0.5em}
\caption{
WikiText-2 perplexity under varying weight and activation quantization bit-widths for models trained on 100B tokens. Three configurations are compared: standard Adam, Muon, and the {OSP} framework.
}
\vspace*{-1em}
\label{fig:100B-quant-plot}
\end{figure}

To systematically evaluate the contribution of each component within the {OSP} framework, we conduct comprehensive ablation studies on a subset of the training corpus comprising 100 billion tokens. This controlled experimental setting enables detailed analysis of how various architectural and optimization choices impact quantization robustness while maintaining computational feasibility.

We assess quantization robustness using two complementary approaches: round-to-nearest (RTN) quantization, and online Hadamard transformations within feed-forward network layers \citep{chee2023quip, liu2024spinquant} that rotate hidden states to address inherent quantization challenges.
Each experimental configuration undergoes evaluation across four quantization scenarios
(Table~\ref{tab:ablation-100B-design-choices}).

The results presented in Table~\ref{tab:ablation-100B-design-choices} demonstrate that
our pre-training strategy
achieves the highest quantization resilience. This configuration attains near-zero excess kurtosis in activation distributions and maintains model performance under both RTN and Hadamard rotation-based quantization. The activation histograms displayed in Figure~\ref{fig:act-hist-100B-comp} particularly highlight the effectiveness of our approach, and the training dynamics illustrated in Figure~\ref{fig:100B-training-log} show that the {OSP} framework uniquely maintains low kurtosis throughout training, providing evidence that our methodology fundamentally eliminates outliers rather than merely reducing them.

Table~\ref{tab:ablation-100B-design-choices} also presents ablation studies that examine the individual contributions of {OSP} components: the Muon optimizer, {\textsc{SSNorm}}, and {\textsc{EmbProj}}.
When using the Muon optimizer as the baseline, switching to the Adam optimizer results in a further increase in kurtosis.
While both {\textsc{SSNorm}} and \mbox{{\textsc{EmbProj}}} independently reduce kurtosis substantially
 compared to Muon alone, the training logs in Figure~\ref{fig:100B-training-log} indicate that neither component using Muon alone is sufficient to prevent
 outlier emergence. This observation suggests that once outliers begin to form at any point in the network, they propagate throughout the entire architecture, ultimately degrading quantization performance.

Figure~\ref{fig:100B-quant-plot} provides comprehensive visualization of perplexity degradation on WikiText-2 across various quantization bit-widths. Our proposed method consistently preserves model performance across all quantization levels, with particularly exceptional performance in the challenging 4-bit regime. Notably, the stability of the performance degradation curve for weights down to 3-bit quantization suggests that our approach fundamentally transforms the model's quantization characteristics rather than merely shifting the performance degradation baseline.


\subsection{Scaling Up to 1 Trillion Tokens}
\begin{table*}[t]
\centering
\resizebox{1.0\textwidth}{!}{%
\begin{tabular}{l|cc|cccccccccc|c}
\toprule
\textbf{Model} & \textbf{Params.} & \textbf{Tokens} & \textbf{ARC} & \textbf{CSQA} & \textbf{GSM8K} & \textbf{HS} & \textbf{MMLU} & \textbf{OBQA} & \textbf{PIQA} & \textbf{SIQA} & \textbf{TQA} & \textbf{WG} & \textbf{Avg.} \\

\midrule
Pythia\nocite{pythia} & 1.4B & 0.3T & 27.2 & 21.5 & 0.0 & 25.8 & 26.2 & 24.8 & 53.2 & 37.2 & 0.0 & 49.0 & 26.5 \\

TinyLlama\nocite{zhang2024tinyllama} & 1.1B & 2T & 28.3 & 22.9 & 0.0 & 26.6 & 26.2 & 21.2 & 48.7 & 40.7 & 0.0 & 49.0 & 26.4 \\

OPT\nocite{zhang2022opt} & 1.3B & 0.3T & 25.0 & 21.6 & 0.0 & 26.5 & 25.6 & 28.2 & 49.6 & 36.9 & 0.0 & 49.5 & 26.3 \\

OLMo\nocite{groeneveld2024olmo} & 1.2B & 3T & 27.7 & 25.8 & 0.0 & 27.0 & 26.1 & 25.8 & 54.1 & 37.4 & 0.0 & \textbf{51.9} & 27.6 \\

Mobile\textsc{LLaMA}\nocite{chu2023mobilellama} & 1.4B & 1.3T & 27.4 & 23.5 & 0.0 & 26.7 & 26.0 & 22.4 & 49.6 & 38.3 & 0.0 & 49.6 & 26.4 \\

Qwen 1.5\nocite{qwen1.5} & 1.8B & 2.4T & 27.2 & 25.4 & 0.0 & 28.3 & 25.7 & 25.0 & 54.1 & 39.1 & 0.0 & 49.3 & 27.4 \\

Qwen 2\nocite{qwen2} & 1.5B & 7T & 30.9 & 27.7 & 0.4 & 35.7 & 26.2 & 28.4 & 56.5 & 38.3 & 0.8 & 48.6 & 29.3 \\

Qwen 2.5\nocite{qwen2.5} & 1.5B & -- & 27.7 & 25.0 & 0.0 & 26.9 & 25.7 & 24.0 & 52.2 & 38.4 & 0.0 & 47.5 & 26.7 \\

\textsc{LLaMA} 3.2\nocite{dubey2024llama} & 1.2B & -- & 29.3 & 24.7 & \textbf{0.5} & 30.1 & 25.8 & 27.4 & 53.3 & 39.5 & 0.1 & 50.3 & 28.1 \\

Stable LM 2\nocite{bellagente2024stablelm2} & 1.6B & 2T & 26.2 & 24.0 & 0.0 & 27.0 & 24.6 & 27.0 & 51.1 & 37.8 & 0.0 & 51.1 & 26.9 \\

SmolLM\nocite{allal2024SmolLM} & 1.7B & 1T & 28.4 & 25.7 & 0.0 & 27.0 & 26.1 & 28.0 & 51.0 & 38.8 & 0.0 & 48.4 & 27.3 \\

SmolLM 2\nocite{allal2025smollm2} & 1.7B & 11T & 25.8 & 22.4 & 0.0 & 25.9 & 24.2 & 26.6 & 51.5 & 36.0 & 0.0 & 50.0 & 26.2 \\

\midrule
\multicolumn{14}{c}{\textbf{From Scratch}} \\
\midrule

Adam & 1.4B & 1T & 25.7 & 25.3 & 0.0 & 26.8 & 25.4 & 26.0 & 49.0 & 37.2 & 0.0 & 49.9 & 26.5 \\

Muon ({OSP}) & 1.4B & 1T & \textbf{45.9} & \textbf{36.2} & \textbf{0.5} & \textbf{44.9} & \textbf{31.1} & \textbf{34.0} & \textbf{65.6} & \textbf{41.3} & \textbf{7.8} & 49.8 & 
\textbf{35.7} \\

\bottomrule
\end{tabular}
}
\caption{
Performance evaluation of 4-bit quantization across 12 open-source large language models and our two models trained from scratch, assessed across 10 benchmark tasks. Model parameters (\textbf{Params.}) represent total trainable parameters, while \textbf{Tokens} indicate training dataset size. Evaluation benchmarks include CommonsenseQA (CSQA), HellaSwag (HS), OpenBookQA (OBQA), TriviaQA (TQA), and WinoGrande (WG), among others. Results demonstrate the impact of extreme quantization on model performance, with the model trained using the {OSP} framework exhibiting superior quantization robustness across the comprehensive benchmark suite.
}
\vspace*{-0.5em}
\label{tab:result-on-ollms-w4a4}
\end{table*}

Having demonstrated the preliminary effectiveness of the {OSP} framework, we scale training to one trillion tokens, matching the scale commonly employed in production environments. While our 100-billion token experiments have confirmed the absence of outlier emergence, model behavior at substantially larger scales remains uncertain. Therefore, validating the effectiveness of each component at production scale constitutes a critical step in verifying the robustness of our framework.

Consistent with our preliminary findings, the kurtosis trajectory demonstrates that the {OSP} framework successfully prevents divergence throughout the entire training process. This result is particularly significant, as
it confirms {OSP}'s effectiveness at the corpus scale in actual production deployments. Unlike previous research limited to preliminary experimental scales, our {OSP} framework provides practical guidelines capable of preventing outlier formation at production scale. Further details are provided in Figure~\ref{fig:1T-training-log}.

Subsequently, Table~\ref{tab:result-on-ollms-w4a4} presents comprehensive benchmark results for 4-bit quantization across 12 open-source LLMs of comparable scale. The majority of baseline models experience severe accuracy degradation under low-bit quantization, with performance on multiple-choice benchmarks such as ARC and CommonsenseQA deteriorating to near-random baselines (25\%). In contrast, our approach demonstrates substantially stronger performance retention, indicating that our framework preserve quantization resilience more effectively.


\section{Analysis}
\subsection{Complementary Benefits with Post-Training Quantization}
\begin{table}
\centering
\begin{tabular}{lrr}

\toprule
\textbf{Quantization} & \textbf{Adam} & \textbf{Muon (OSP)} \\
\midrule
RTN & 14475.51 & \textbf{45.92} \\
+ $\text{FFN Had}^\dagger$ & 4794.00 & \textbf{19.27} \\
+ GPTQ & 3723.46 & \textbf{14.29} \\
+ QuaRot & 16.62 & \textbf{14.38} \\
+ SpinQuant & 14.94 & \textbf{13.66} \\
\bottomrule

\end{tabular}
\caption{
WikiText-2 perplexity after applying various PTQ methods under 4-bit quantization. Minimal methods (Hadamard and GPTQ) show limited effectiveness on Adam. Models trained with {OSP} demonstrate consistently superior performance across all scenarios. $^\dagger$Only applies Hadamard transform to FFN hidden states.
}
\vspace*{-1em}
\label{tab:with-ptq}
\end{table}

To further investigate the practical implications of our framework, we examine how models trained with {OSP} perform when combined with existing PTQ techniques. Unlike architectural modifications that alter model structure, our approach maintains computational invariance while exhibiting fundamentally different quantization characteristics. This raises an important question: does the absence of outliers eliminate the need for PTQ?

Table~\ref{tab:with-ptq} demonstrates that our model trained under {OSP} achieves complementary performance gains when combined with PTQ methods. This finding is particularly significant because, unlike Quantization-Aware Training (QAT), our framework does not incorporate explicit quantization objectives during pre-training. Rather, our primary goal centers on eliminating the outlier features that fundamentally impede quantization, thereby creating models with inherently superior quantization robustness. While our framework demonstrates consistent perplexity improvements over RTN quantization across various PTQ scenarios, the gains are less pronounced than those achieved with Adam. Crucially, however, our approach consistently outperforms Adam across all evaluation settings. This pattern suggests that our approach provides a better foundation for subsequent PTQ calibration.

\subsection{Attention Sinks without Outliers}

\begin{figure}[t]
\centering

\hspace{-1.5em}
\begin{minipage}[b]{0.48\linewidth}
    \begin{flushleft}
        \includegraphics[width=1.15\linewidth]{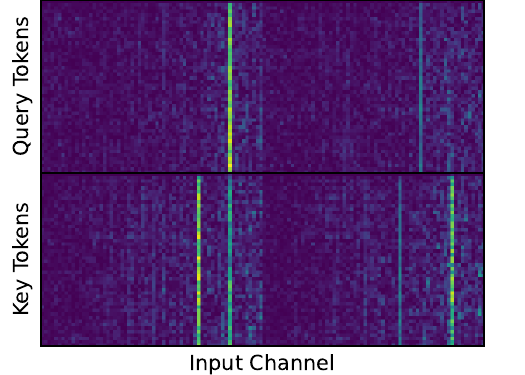}
    \end{flushleft}
  \hspace{0.5em}
  \centerline{\textbf{(a) Adam}}
\end{minipage}
\hspace{0.5em}
\begin{minipage}[b]{0.48\linewidth}
    \begin{flushleft}
        \includegraphics[width=1.15\linewidth]{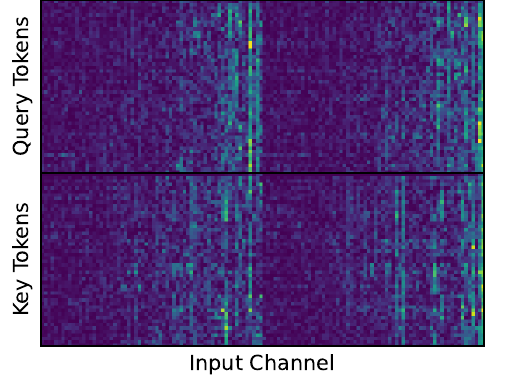}
    \end{flushleft}
  \centerline{\textbf{(b) Muon (OSP)}}
\end{minipage}

\caption{
Activation magnitudes of query and key tokens within attention sink heads comparing Adam and OSP models. Adam models exhibit concentrated activation patterns with sparse high-magnitude channels, while OSP demonstrates broadly distributed activation patterns across multiple channels.
}
\vspace*{-1em}
\label{fig:attn-qk-mag}
\end{figure}
\begin{figure}[t]
\centering
\includegraphics[width=1\linewidth]{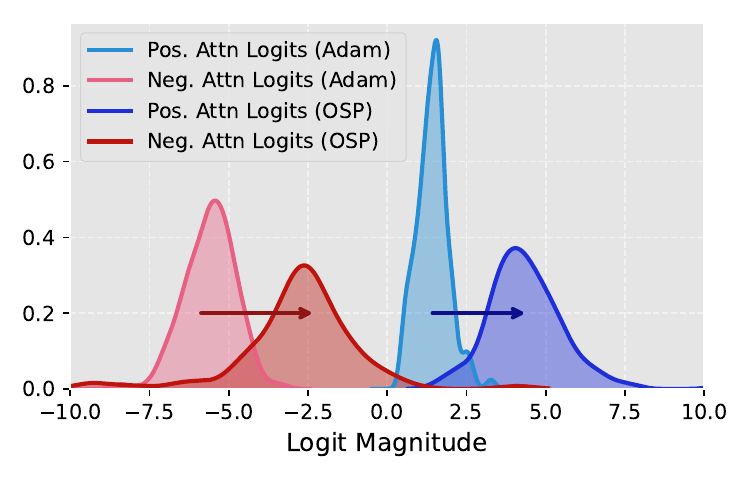}
\vspace*{-2em}
\caption{
Attention logit distributions at sink token positions comparing Adam and OSP training. Adam models exhibit skewed distributions with predominantly negative logits, while OSP models demonstrate balanced distributions between sink tokens and other positions. 
}
\vspace*{-1em}
\label{fig:attn-sink-logit-dist}
\end{figure}

We conduct a qualitative analysis to examine the internal dynamics of models when outliers are eliminated. Following previous studies that have conceptualized outliers as byproducts of attention sinks, we investigate whether the attention sink phenomenon disappears when massive activations are mitigated.

Following \citet{bondarenko2021understanding}, we initially identify activations that exceed 6 standard deviations from the mean. As a result, models trained with Adam exhibit massive activations concentrated in delimiter tokens and similar positions, consistent with prior findings. In contrast, our framework demonstrates complete absence of such massive activations across all examined components.

Despite the elimination of massive activations, attention sink patterns still persist in our models. Following the analytical approach of \citet{gu2025attnsinkemerges}, we apply threshold-based filtering to identify cases where attention concentrates heavily on initial tokens. This analysis confirms that OSP-trained models continue to exhibit substantial attention sink behavior, raising important questions about the underlying mechanisms linking attention patterns to outlier formation.

To address this apparent contradiction, we conduct a preliminary analysis by examining the distributional characteristics of attention mechanisms in both training paradigms. Figure~\ref{fig:attn-qk-mag} visualizes the query and key activation magnitudes within attention heads that exhibit sink behavior. Models trained with Adam demonstrate a concentration of massive values within a small number of outlier channels, while the OSP-trained model distributes these magnitudes broadly across multiple channels.

This distributional difference is directly related to the attention logit patterns illustrated in Figure~\ref{fig:attn-sink-logit-dist}, which compares the logit distributions between sink tokens and other sequence positions. For models prone to massive activations, the concentration of large values in overlapping channels generates predominantly negative logits across most tokens.
Conversely, OSP-trained models achieve more uniform distributions, resulting in balanced logits between sink tokens and the broader token sequence.

Since softmax normalization operates on relative logit differences rather than absolute values, models can achieve effective "no-op" operations without driving attention logits toward negative infinity. Our analysis suggests that attention sinks do not inherently cause massive activations. Rather, models prone to outlier formation tend to adopt the negative infinity strategy as a computational solution for implementing "no-op" operations within training dynamics that favor concentrated channel activations.

\section{Conclusion}

We present the Outlier-Safe Pre-Training (OSP) framework, which prevents emergence of activation outliers during LLM training by replacing Adam with the Muon optimizer, adopting Single-Scale RMSNorm, and incorporating learnable embedding projections. Our approach achieves production-scale efficiency by training a 1.4B model on 1 trillion tokens with only 2\% overhead compared to Adam, while fundamentally improving quantization robustness. The resulting model maintains strong performance under aggressive 4-bit quantization where comparable models fail catastrophically. By preventing outliers instead of mitigating them post-hoc, OSP enables robust low-bit deployment without architectural modifications at inference time or costly quantization-aware training. Our publicly released model provides the first demonstration that outlier-free training is both feasible and practical at scale, opening new possibilities for efficient LLM deployment.
\section*{Limitations}
Our study focuses primarily on Muon without extensive comparisons to other second-order methods like Shampoo or SOAP. This limitation stems from practical constraints: TPU compilation time for training pipelines often exceeds one hour, making comprehensive optimizer ablation studies prohibitively time-consuming given our available computational resources.

Additionally, while our experiments demonstrate effectiveness on a 1.4B-parameter model, we have not yet explored the impact across a range of model sizes, particularly the 3B and 7B parameter scales commonly targeted for mobile deployment. Looking ahead, we plan to extend our analysis to these larger models. Our distributed implementation of Muon in JAX achieves comparable efficiency to Adam, making such broader experiments computationally feasible.
%

\subsection*{Acknowledgments}
This work was supported in part by
the National Research Foundation of Korea [NRF-2023R1A2C3004176, RS-2023-00262002],
the Ministry of SMEs and Startups [RS-2024-00523644],
the Ministry of Health \& Welfare, Republic of Korea [HR20C002103],
the ICT Creative Consilience program through the Institute of Information \& Communications Technology Planning \& Evaluation (IITP) grant funded by the MSIT [IITP-2025-RS-2020-II201819],
and Cloud TPUs from Google’s TPU Research Cloud (TRC).

\bibliography{custom}

\begin{thebibliography}{68}
\providecommand{\natexlab}[1]{#1}

\bibitem[{Allal et~al.(2025)Allal, Lozhkov, Bakouch, Blázquez, Penedo, Tunstall, Marafioti, Kydlíček, Lajarín, Srivastav, Lochner, Fahlgren, Nguyen, Fourrier, Burtenshaw, Larcher, Zhao, Zakka, Morlon, Raffel, von Werra, and Wolf}]{allal2025smollm2}
Loubna~Ben Allal, Anton Lozhkov, Elie Bakouch, Gabriel~Martín Blázquez, Guilherme Penedo, Lewis Tunstall, Andrés Marafioti, Hynek Kydlíček, Agustín~Piqueres Lajarín, Vaibhav Srivastav, Joshua Lochner, Caleb Fahlgren, Xuan-Son Nguyen, Clémentine Fourrier, Ben Burtenshaw, Hugo Larcher, Haojun Zhao, Cyril Zakka, Mathieu Morlon, Colin Raffel, Leandro von Werra, and Thomas Wolf. 2025.
\newblock \href {https://arxiv.org/abs/2502.02737} {Smollm2: When smol goes big -- data-centric training of a small language model}.
\newblock \emph{Preprint}, arXiv:2502.02737.

\bibitem[{Allal et~al.(2024)Allal, Lozhkov, Bakouch, von Werra, and Wolf}]{allal2024SmolLM}
Loubna~Ben Allal, Anton Lozhkov, Elie Bakouch, Leandro von Werra, and Thomas Wolf. 2024.
\newblock Smollm - blazingly fast and remarkably powerful.

\bibitem[{Ashkboos et~al.(2024{\natexlab{a}})Ashkboos, Croci, do~Nascimento, Hoefler, and Hensman}]{ashkboos2024slicegpt}
Saleh Ashkboos, Maximilian~L. Croci, Marcelo~Gennari do~Nascimento, Torsten Hoefler, and James Hensman. 2024{\natexlab{a}}.
\newblock \href {https://openreview.net/forum?id=vXxardq6db} {Slice{GPT}: Compress large language models by deleting rows and columns}.
\newblock In \emph{The Twelfth International Conference on Learning Representations}.

\bibitem[{Ashkboos et~al.(2024{\natexlab{b}})Ashkboos, Mohtashami, Croci, Li, Cameron, Jaggi, Alistarh, Hoefler, and Hensman}]{ashkboos2024quarot}
Saleh Ashkboos, Amirkeivan Mohtashami, Maximilian Croci, Bo~Li, Pashmina Cameron, Martin Jaggi, Dan Alistarh, Torsten Hoefler, and James Hensman. 2024{\natexlab{b}}.
\newblock \href {https://proceedings.neurips.cc/paper_files/paper/2024/file/b5b939436789f76f08b9d0da5e81af7c-Paper-Conference.pdf} {Quarot: Outlier-free 4-bit inference in rotated llms}.
\newblock In \emph{Advances in Neural Information Processing Systems}, volume~37, pages 100213--100240. Curran Associates, Inc.

\bibitem[{Bai et~al.(2023)Bai, Bai, Chu, Cui, Dang, Deng, Fan, Ge, Han, Huang, Hui, Ji, Li, Lin, Lin, Liu, Liu, Lu, Lu, Ma, Men, Ren, Ren, Tan, Tan, Tu, Wang, Wang, Wang, Wu, Xu, Xu, Yang, Yang, Yang, Yang, Yao, Yu, Yuan, Yuan, Zhang, Zhang, Zhang, Zhang, Zhou, Zhou, Zhou, and Zhu}]{qwen1.5}
Jinze Bai, Shuai Bai, Yunfei Chu, Zeyu Cui, Kai Dang, Xiaodong Deng, Yang Fan, Wenbin Ge, Yu~Han, Fei Huang, Binyuan Hui, Luo Ji, Mei Li, Junyang Lin, Runji Lin, Dayiheng Liu, Gao Liu, Chengqiang Lu, Keming Lu, Jianxin Ma, Rui Men, Xingzhang Ren, Xuancheng Ren, Chuanqi Tan, Sinan Tan, Jianhong Tu, Peng Wang, Shijie Wang, Wei Wang, Shengguang Wu, Benfeng Xu, Jin Xu, An~Yang, Hao Yang, Jian Yang, Shusheng Yang, Yang Yao, Bowen Yu, Hongyi Yuan, Zheng Yuan, Jianwei Zhang, Xingxuan Zhang, Yichang Zhang, Zhenru Zhang, Chang Zhou, Jingren Zhou, Xiaohuan Zhou, and Tianhang Zhu. 2023.
\newblock Qwen technical report.
\newblock \emph{arXiv preprint arXiv:2309.16609}.

\bibitem[{Barbero et~al.(2025)Barbero, Arroyo, Gu, Perivolaropoulos, Bronstein, Veli{\v{c}}kovi{\'c}, and Pascanu}]{barbero2025whyllmsattendfirsttoken}
Federico Barbero, Alvaro Arroyo, Xiangming Gu, Christos Perivolaropoulos, Michael Bronstein, Petar Veli{\v{c}}kovi{\'c}, and Razvan Pascanu. 2025.
\newblock Why do llms attend to the first token?
\newblock \emph{arXiv preprint arXiv:2504.02732}.

\bibitem[{Bellagente et~al.(2024)Bellagente, Tow, Mahan, Phung, Zhuravinskyi, Adithyan, Baicoianu, Brooks, Cooper, Datta et~al.}]{bellagente2024stablelm2}
Marco Bellagente, Jonathan Tow, Dakota Mahan, Duy Phung, Maksym Zhuravinskyi, Reshinth Adithyan, James Baicoianu, Ben Brooks, Nathan Cooper, Ashish Datta, et~al. 2024.
\newblock Stable lm 2 1.6 b technical report.
\newblock \emph{arXiv preprint arXiv:2402.17834}.

\bibitem[{Ben~Allal et~al.(2024)Ben~Allal, Lozhkov, Penedo, Wolf, and von Werra}]{benallal2024cosmopedia}
Loubna Ben~Allal, Anton Lozhkov, Guilherme Penedo, Thomas Wolf, and Leandro von Werra. 2024.
\newblock \href {https://huggingface.co/datasets/HuggingFaceTB/cosmopedia} {Cosmopedia}.

\bibitem[{Bernstein and Newhouse(2024{\natexlab{a}})}]{bernstein2024modularduality}
Jeremy Bernstein and Laker Newhouse. 2024{\natexlab{a}}.
\newblock Modular duality in deep learning.
\newblock \emph{arXiv preprint arXiv:2410.21265}.

\bibitem[{Bernstein and Newhouse(2024{\natexlab{b}})}]{bernstein2024oldoptnewnorm}
Jeremy Bernstein and Laker Newhouse. 2024{\natexlab{b}}.
\newblock Old optimizer, new norm: An anthology.
\newblock \emph{arXiv preprint arXiv:2409.20325}.

\bibitem[{Biderman et~al.(2023)Biderman, Schoelkopf, Anthony, Bradley, O'Brien, Hallahan, Khan, Purohit, Prashanth, Raff, Skowron, Sutawika, and Van Der~Wal}]{pythia}
Stella Biderman, Hailey Schoelkopf, Quentin~Gregory Anthony, Herbie Bradley, Kyle O'Brien, Eric Hallahan, Mohammad~Aflah Khan, Shivanshu Purohit, Usvsn~Sai Prashanth, Edward Raff, Aviya Skowron, Lintang Sutawika, and Oskar Van Der~Wal. 2023.
\newblock \href {https://proceedings.mlr.press/v202/biderman23a.html} {Pythia: A suite for analyzing large language models across training and scaling}.
\newblock In \emph{Proceedings of the 40th International Conference on Machine Learning}, volume 202 of \emph{Proceedings of Machine Learning Research}, pages 2397--2430. PMLR.

\bibitem[{Bisk et~al.(2020)Bisk, Zellers, Le~bras, Gao, and Choi}]{Bisk2020piqa}
Yonatan Bisk, Rowan Zellers, Ronan Le~bras, Jianfeng Gao, and Yejin Choi. 2020.
\newblock \href {https://doi.org/10.1609/aaai.v34i05.6239} {Piqa: Reasoning about physical commonsense in natural language}.
\newblock \emph{Proceedings of the AAAI Conference on Artificial Intelligence}, 34(05):7432--7439.

\bibitem[{Bondarenko et~al.(2021)Bondarenko, Nagel, and Blankevoort}]{bondarenko2021understanding}
Yelysei Bondarenko, Markus Nagel, and Tijmen Blankevoort. 2021.
\newblock \href {https://doi.org/10.18653/v1/2021.emnlp-main.627} {Understanding and overcoming the challenges of efficient transformer quantization}.
\newblock In \emph{Proceedings of the 2021 Conference on Empirical Methods in Natural Language Processing}, pages 7947--7969, Online and Punta Cana, Dominican Republic. Association for Computational Linguistics.

\bibitem[{Bondarenko et~al.(2023)Bondarenko, Nagel, and Blankevoort}]{bondarenko2023quantizabletransformers}
Yelysei Bondarenko, Markus Nagel, and Tijmen Blankevoort. 2023.
\newblock \href {https://proceedings.neurips.cc/paper_files/paper/2023/file/edbcb7583fd8921dad78adecfe06a99b-Paper-Conference.pdf} {Quantizable transformers: Removing outliers by helping attention heads do nothing}.
\newblock In \emph{Advances in Neural Information Processing Systems}, volume~36, pages 75067--75096. Curran Associates, Inc.

\bibitem[{Bradbury et~al.(2018)Bradbury, Frostig, Hawkins, Johnson, Leary, Maclaurin, Necula, Paszke, Vander{P}las, Wanderman-{M}ilne, and Zhang}]{jax2018github}
James Bradbury, Roy Frostig, Peter Hawkins, Matthew~James Johnson, Chris Leary, Dougal Maclaurin, George Necula, Adam Paszke, Jake Vander{P}las, Skye Wanderman-{M}ilne, and Qiao Zhang. 2018.
\newblock \href {http://github.com/jax-ml/jax} {{JAX}: composable transformations of {P}ython+{N}um{P}y programs}.

\bibitem[{Caples and Neuhaus(2024)}]{caples2024adamprivileged}
Diego Caples and Rob Neuhaus. 2024.
\newblock \href {https://www.lesswrong.com/posts/yrhu6MeFddnGRSLtQ/adam-optimizer-causes-privileged-basis-in-transformer-lm} {Adam optimizer causes privileged basis in transformer lm residual stream}.
\newblock \emph{LessWrong}.

\bibitem[{Chee et~al.(2023)Chee, Cai, Kuleshov, and De~Sa}]{chee2023quip}
Jerry Chee, Yaohui Cai, Volodymyr Kuleshov, and Christopher~M De~Sa. 2023.
\newblock \href {https://proceedings.neurips.cc/paper_files/paper/2023/file/0df38cd13520747e1e64e5b123a78ef8-Paper-Conference.pdf} {Quip: 2-bit quantization of large language models with guarantees}.
\newblock In \emph{Advances in Neural Information Processing Systems}, volume~36, pages 4396--4429. Curran Associates, Inc.

\bibitem[{Chen et~al.(2024)Chen, Shao, Xu, Wang, Gao, Zhang, Qiao, and Luo}]{efficientqat}
Mengzhao Chen, Wenqi Shao, Peng Xu, Jiahao Wang, Peng Gao, Kaipeng Zhang, Yu~Qiao, and Ping Luo. 2024.
\newblock Efficientqat: Efficient quantization-aware training for large language models.
\newblock \emph{arXiv preprint arXiv:2407.11062}.

\bibitem[{Chu et~al.(2023)Chu, Qiao, Lin, Xu, Yang, Hu, Wei, Zhang, Zhang, Wei et~al.}]{chu2023mobilellama}
Xiangxiang Chu, Limeng Qiao, Xinyang Lin, Shuang Xu, Yang Yang, Yiming Hu, Fei Wei, Xinyu Zhang, Bo~Zhang, Xiaolin Wei, et~al. 2023.
\newblock Mobilevlm: A fast, reproducible and strong vision language assistant for mobile devices.
\newblock \emph{arXiv preprint arXiv:2312.16886}.

\bibitem[{Clark et~al.(2018)Clark, Cowhey, Etzioni, Khot, Sabharwal, Schoenick, and Tafjord}]{allenai:arc}
Peter Clark, Isaac Cowhey, Oren Etzioni, Tushar Khot, Ashish Sabharwal, Carissa Schoenick, and Oyvind Tafjord. 2018.
\newblock Think you have solved question answering? try arc, the ai2 reasoning challenge.
\newblock \emph{arXiv:1803.05457v1}.

\bibitem[{Cobbe et~al.(2021)Cobbe, Kosaraju, Bavarian, Chen, Jun, Kaiser, Plappert, Tworek, Hilton, Nakano, Hesse, and Schulman}]{cobbe2021gsm8k}
Karl Cobbe, Vineet Kosaraju, Mohammad Bavarian, Mark Chen, Heewoo Jun, Lukasz Kaiser, Matthias Plappert, Jerry Tworek, Jacob Hilton, Reiichiro Nakano, Christopher Hesse, and John Schulman. 2021.
\newblock Training verifiers to solve math word problems.
\newblock \emph{arXiv preprint arXiv:2110.14168}.

\bibitem[{Dettmers et~al.(2022)Dettmers, Lewis, Belkada, and Zettlemoyer}]{dettmers2022llmint8}
Tim Dettmers, Mike Lewis, Younes Belkada, and Luke Zettlemoyer. 2022.
\newblock \href {https://proceedings.neurips.cc/paper_files/paper/2022/file/c3ba4962c05c49636d4c6206a97e9c8a-Paper-Conference.pdf} {Gpt3.int8(): 8-bit matrix multiplication for transformers at scale}.
\newblock In \emph{Advances in Neural Information Processing Systems}, volume~35, pages 30318--30332. Curran Associates, Inc.

\bibitem[{Dubey et~al.(2024)Dubey, Jauhri, Pandey, Kadian, Al-Dahle, Letman, Mathur, Schelten, Yang, Fan et~al.}]{dubey2024llama}
Abhimanyu Dubey, Abhinav Jauhri, Abhinav Pandey, Abhishek Kadian, Ahmad Al-Dahle, Aiesha Letman, Akhil Mathur, Alan Schelten, Amy Yang, Angela Fan, et~al. 2024.
\newblock The llama 3 herd of models.
\newblock \emph{arXiv preprint arXiv:2407.21783}.

\bibitem[{Duvvuri et~al.(2024)Duvvuri, Devvrit, Anil, Hsieh, and Dhillon}]{duvvuri2024caspr}
Sai~Surya Duvvuri, Fnu Devvrit, Rohan Anil, Cho-Jui Hsieh, and Inderjit~S Dhillon. 2024.
\newblock \href {https://openreview.net/forum?id=8j9hz8DVi8} {Combining axes preconditioners through kronecker approximation for deep learning}.
\newblock In \emph{The Twelfth International Conference on Learning Representations}.

\bibitem[{Elhage et~al.(2023)Elhage, Lasenby, and Olah}]{elhage2023privilegedbases}
Nelson Elhage, Robert Lasenby, and Christopher Olah. 2023.
\newblock \href {https://transformer-circuits.pub/2023/privileged-basis/index.html} {Privileged bases in the transformer residual stream}.
\newblock \emph{Transformer Circuits Thread}.

\bibitem[{Frantar et~al.(2023)Frantar, Ashkboos, Hoefler, and Alistarh}]{frantar-gptq}
Elias Frantar, Saleh Ashkboos, Torsten Hoefler, and Dan Alistarh. 2023.
\newblock \href {https://openreview.net/forum?id=tcbBPnfwxS} {{OPTQ}: Accurate quantization for generative pre-trained transformers}.
\newblock In \emph{The Eleventh International Conference on Learning Representations}.

\bibitem[{Groeneveld et~al.(2024)Groeneveld, Beltagy, Walsh, Bhagia, Kinney, Tafjord, Jha, Ivison, Magnusson, Wang, Arora, Atkinson, Authur, Chandu, Cohan, Dumas, Elazar, Gu, Hessel, Khot, Merrill, Morrison, Muennighoff, Naik, Nam, Peters, Pyatkin, Ravichander, Schwenk, Shah, Smith, Strubell, Subramani, Wortsman, Dasigi, Lambert, Richardson, Zettlemoyer, Dodge, Lo, Soldaini, Smith, and Hajishirzi}]{groeneveld2024olmo}
Dirk Groeneveld, Iz~Beltagy, Evan Walsh, Akshita Bhagia, Rodney Kinney, Oyvind Tafjord, Ananya Jha, Hamish Ivison, Ian Magnusson, Yizhong Wang, Shane Arora, David Atkinson, Russell Authur, Khyathi Chandu, Arman Cohan, Jennifer Dumas, Yanai Elazar, Yuling Gu, Jack Hessel, Tushar Khot, William Merrill, Jacob Morrison, Niklas Muennighoff, Aakanksha Naik, Crystal Nam, Matthew Peters, Valentina Pyatkin, Abhilasha Ravichander, Dustin Schwenk, Saurabh Shah, William Smith, Emma Strubell, Nishant Subramani, Mitchell Wortsman, Pradeep Dasigi, Nathan Lambert, Kyle Richardson, Luke Zettlemoyer, Jesse Dodge, Kyle Lo, Luca Soldaini, Noah Smith, and Hannaneh Hajishirzi. 2024.
\newblock \href {https://doi.org/10.18653/v1/2024.acl-long.841} {{OLM}o: Accelerating the science of language models}.
\newblock In \emph{Proceedings of the 62nd Annual Meeting of the Association for Computational Linguistics (Volume 1: Long Papers)}, pages 15789--15809, Bangkok, Thailand. Association for Computational Linguistics.

\bibitem[{Gu et~al.(2025)Gu, Pang, Du, Liu, Zhang, Du, Wang, and Lin}]{gu2025attnsinkemerges}
Xiangming Gu, Tianyu Pang, Chao Du, Qian Liu, Fengzhuo Zhang, Cunxiao Du, Ye~Wang, and Min Lin. 2025.
\newblock \href {https://openreview.net/forum?id=78Nn4QJTEN} {When attention sink emerges in language models: An empirical view}.
\newblock In \emph{The Thirteenth International Conference on Learning Representations}.

\bibitem[{Guo et~al.(2024)Guo, Pai, Bai, Jiao, Jordan, and Mei}]{guo2024activedormantattnheads}
Tianyu Guo, Druv Pai, Yu~Bai, Jiantao Jiao, Michael~I Jordan, and Song Mei. 2024.
\newblock Active-dormant attention heads: Mechanistically demystifying extreme-token phenomena in llms.
\newblock \emph{arXiv preprint arXiv:2410.13835}.

\bibitem[{Gupta et~al.(2018)Gupta, Koren, and Singer}]{gupta2018shampoo}
Vineet Gupta, Tomer Koren, and Yoram Singer. 2018.
\newblock \href {https://proceedings.mlr.press/v80/gupta18a.html} {Shampoo: Preconditioned stochastic tensor optimization}.
\newblock In \emph{Proceedings of the 35th International Conference on Machine Learning}, volume~80 of \emph{Proceedings of Machine Learning Research}, pages 1842--1850. PMLR.

\bibitem[{H\"{a}gele et~al.(2024)H\"{a}gele, Bakouch, Kosson, Ben~allal, Von~Werra, and Jaggi}]{hagele2024trapezoidal}
Alex H\"{a}gele, Elie Bakouch, Atli Kosson, Loubna Ben~allal, Leandro Von~Werra, and Martin Jaggi. 2024.
\newblock \href {https://proceedings.neurips.cc/paper_files/paper/2024/file/8b970e15a89bf5d12542810df8eae8fc-Paper-Conference.pdf} {Scaling laws and compute-optimal training beyond fixed training durations}.
\newblock In \emph{Advances in Neural Information Processing Systems}, volume~37, pages 76232--76264. Curran Associates, Inc.

\bibitem[{He et~al.(2024)He, Noci, Paliotta, Schlag, and Hofmann}]{he2024OP}
Bobby He, Lorenzo Noci, Daniele Paliotta, Imanol Schlag, and Thomas Hofmann. 2024.
\newblock \href {https://proceedings.neurips.cc/paper_files/paper/2024/file/986292a930c3692168b177a770025ab3-Paper-Conference.pdf} {Understanding and minimising outlier features in transformer training}.
\newblock In \emph{Advances in Neural Information Processing Systems}, volume~37, pages 83786--83846. Curran Associates, Inc.

\bibitem[{Hendrycks et~al.(2021)Hendrycks, Burns, Basart, Zou, Mazeika, Song, and Steinhardt}]{hendryckstest2021mmlu}
Dan Hendrycks, Collin Burns, Steven Basart, Andy Zou, Mantas Mazeika, Dawn Song, and Jacob Steinhardt. 2021.
\newblock Measuring massive multitask language understanding.
\newblock \emph{Proceedings of the International Conference on Learning Representations (ICLR)}.

\bibitem[{Higham(2008)}]{higham2008newtonschulz}
NJ~Higham. 2008.
\newblock Functions of matrices: Theory and computation.

\bibitem[{Jordan et~al.(2024)Jordan, Jin, Boza, Jiacheng, Cecista, Newhouse, and Bernstein}]{jordan2024muon}
Keller Jordan, Yuchen Jin, Vlado Boza, You Jiacheng, Franz Cecista, Laker Newhouse, and Jeremy Bernstein. 2024.
\newblock \href {https://kellerjordan.github.io/posts/muon/} {Muon: An optimizer for hidden layers in neural networks}.

\bibitem[{Joshi et~al.(2017)Joshi, Choi, Weld, and Zettlemoyer}]{joshi2017triviaqa}
Mandar Joshi, Eunsol Choi, Daniel Weld, and Luke Zettlemoyer. 2017.
\newblock \href {https://doi.org/10.18653/v1/P17-1147} {{T}rivia{QA}: A large scale distantly supervised challenge dataset for reading comprehension}.
\newblock In \emph{Proceedings of the 55th Annual Meeting of the Association for Computational Linguistics (Volume 1: Long Papers)}, pages 1601--1611, Vancouver, Canada. Association for Computational Linguistics.

\bibitem[{Kim et~al.(2024)Kim, Hooper, Gholami, Dong, Li, Shen, Mahoney, and Keutzer}]{kim2024squeezellm}
Sehoon Kim, Coleman Richard~Charles Hooper, Amir Gholami, Zhen Dong, Xiuyu Li, Sheng Shen, Michael~W. Mahoney, and Kurt Keutzer. 2024.
\newblock \href {https://proceedings.mlr.press/v235/kim24f.html} {{S}queeze{LLM}: Dense-and-sparse quantization}.
\newblock In \emph{Proceedings of the 41st International Conference on Machine Learning}, volume 235 of \emph{Proceedings of Machine Learning Research}, pages 23901--23923. PMLR.

\bibitem[{Kovaleva et~al.(2021)Kovaleva, Kulshreshtha, Rogers, and Rumshisky}]{kovaleva-etal-2021-bert-busters}
Olga Kovaleva, Saurabh Kulshreshtha, Anna Rogers, and Anna Rumshisky. 2021.
\newblock \href {https://doi.org/10.18653/v1/2021.findings-acl.300} {{BERT} busters: Outlier dimensions that disrupt transformers}.
\newblock In \emph{Findings of the Association for Computational Linguistics: ACL-IJCNLP 2021}, pages 3392--3405, Online. Association for Computational Linguistics.

\bibitem[{Li et~al.(2023)Li, Allal, Zi, Muennighoff, Kocetkov, Mou, Marone, Akiki, Li, Chim, Liu, Zheltonozhskii, Zhuo, Wang, Dehaene, Davaadorj, Lamy-Poirier, Monteiro, Shliazhko, Gontier, Meade, Zebaze, Yee, Umapathi, Zhu, Lipkin, Oblokulov, Wang, Murthy, Stillerman, Patel, Abulkhanov, Zocca, Dey, Zhang, Fahmy, Bhattacharyya, Yu, Singh, Luccioni, Villegas, Kunakov, Zhdanov, Romero, Lee, Timor, Ding, Schlesinger, Schoelkopf, Ebert, Dao, Mishra, Gu, Robinson, Anderson, Dolan-Gavitt, Contractor, Reddy, Fried, Bahdanau, Jernite, Ferrandis, Hughes, Wolf, Guha, von Werra, and de~Vries}]{li2023starcoder}
Raymond Li, Loubna~Ben Allal, Yangtian Zi, Niklas Muennighoff, Denis Kocetkov, Chenghao Mou, Marc Marone, Christopher Akiki, Jia Li, Jenny Chim, Qian Liu, Evgenii Zheltonozhskii, Terry~Yue Zhuo, Thomas Wang, Olivier Dehaene, Mishig Davaadorj, Joel Lamy-Poirier, João Monteiro, Oleh Shliazhko, Nicolas Gontier, Nicholas Meade, Armel Zebaze, Ming-Ho Yee, Logesh~Kumar Umapathi, Jian Zhu, Benjamin Lipkin, Muhtasham Oblokulov, Zhiruo Wang, Rudra Murthy, Jason Stillerman, Siva~Sankalp Patel, Dmitry Abulkhanov, Marco Zocca, Manan Dey, Zhihan Zhang, Nour Fahmy, Urvashi Bhattacharyya, Wenhao Yu, Swayam Singh, Sasha Luccioni, Paulo Villegas, Maxim Kunakov, Fedor Zhdanov, Manuel Romero, Tony Lee, Nadav Timor, Jennifer Ding, Claire Schlesinger, Hailey Schoelkopf, Jan Ebert, Tri Dao, Mayank Mishra, Alex Gu, Jennifer Robinson, Carolyn~Jane Anderson, Brendan Dolan-Gavitt, Danish Contractor, Siva Reddy, Daniel Fried, Dzmitry Bahdanau, Yacine Jernite, Carlos~Muñoz Ferrandis, Sean Hughes, Thomas Wolf, Arjun Guha, Leandro von
  Werra, and Harm de~Vries. 2023.
\newblock \href {https://arxiv.org/abs/2305.06161} {Starcoder: may the source be with you!}
\newblock \emph{arXiv preprint arXiv:2305.06161}.

\bibitem[{Lin et~al.(2024)Lin, Tang, Tang, Yang, Chen, Wang, Xiao, Dang, Gan, and Han}]{lin2024awq}
Ji~Lin, Jiaming Tang, Haotian Tang, Shang Yang, Wei-Ming Chen, Wei-Chen Wang, Guangxuan Xiao, Xingyu Dang, Chuang Gan, and Song Han. 2024.
\newblock \href {https://proceedings.mlsys.org/paper_files/paper/2024/file/42a452cbafa9dd64e9ba4aa95cc1ef21-Paper-Conference.pdf} {Awq: Activation-aware weight quantization for on-device llm compression and acceleration}.
\newblock In \emph{Proceedings of Machine Learning and Systems}, volume~6, pages 87--100.

\bibitem[{Liu et~al.(2024{\natexlab{a}})Liu, Bai, Lin, Li, Gao, Xu, Hou, Yao, and Yuan}]{liu-etal-2024-intactkv}
Ruikang Liu, Haoli Bai, Haokun Lin, Yuening Li, Han Gao, Zhengzhuo Xu, Lu~Hou, Jun Yao, and Chun Yuan. 2024{\natexlab{a}}.
\newblock \href {https://doi.org/10.18653/v1/2024.findings-acl.460} {{I}ntact{KV}: Improving large language model quantization by keeping pivot tokens intact}.
\newblock In \emph{Findings of the Association for Computational Linguistics: ACL 2024}, pages 7716--7741, Bangkok, Thailand. Association for Computational Linguistics.

\bibitem[{Liu et~al.(2024{\natexlab{b}})Liu, Oguz, Zhao, Chang, Stock, Mehdad, Shi, Krishnamoorthi, and Chandra}]{liu2024llmqat}
Zechun Liu, Barlas Oguz, Changsheng Zhao, Ernie Chang, Pierre Stock, Yashar Mehdad, Yangyang Shi, Raghuraman Krishnamoorthi, and Vikas Chandra. 2024{\natexlab{b}}.
\newblock \href {https://doi.org/10.18653/v1/2024.findings-acl.26} {{LLM}-{QAT}: Data-free quantization aware training for large language models}.
\newblock In \emph{Findings of the Association for Computational Linguistics: ACL 2024}, pages 467--484, Bangkok, Thailand. Association for Computational Linguistics.

\bibitem[{Liu et~al.(2024{\natexlab{c}})Liu, Zhao, Fedorov, Soran, Choudhary, Krishnamoorthi, Chandra, Tian, and Blankevoort}]{liu2024spinquant}
Zechun Liu, Changsheng Zhao, Igor Fedorov, Bilge Soran, Dhruv Choudhary, Raghuraman Krishnamoorthi, Vikas Chandra, Yuandong Tian, and Tijmen Blankevoort. 2024{\natexlab{c}}.
\newblock Spinquant--llm quantization with learned rotations.
\newblock \emph{arXiv preprint arXiv:2405.16406}.

\bibitem[{Lozhkov et~al.(2024)Lozhkov, Ben~Allal, von Werra, and Wolf}]{lozhkov2024fineweb-edu}
Anton Lozhkov, Loubna Ben~Allal, Leandro von Werra, and Thomas Wolf. 2024.
\newblock \href {https://doi.org/10.57967/hf/2497} {Fineweb-edu: the finest collection of educational content}.

\bibitem[{Martens and Grosse(2015)}]{martens2015kfac}
James Martens and Roger Grosse. 2015.
\newblock \href {https://proceedings.mlr.press/v37/martens15.html} {Optimizing neural networks with kronecker-factored approximate curvature}.
\newblock In \emph{Proceedings of the 32nd International Conference on Machine Learning}, volume~37 of \emph{Proceedings of Machine Learning Research}, pages 2408--2417, Lille, France. PMLR.

\bibitem[{Merity et~al.(2016)Merity, Xiong, Bradbury, and Socher}]{merity2016wikitext2}
Stephen Merity, Caiming Xiong, James Bradbury, and Richard Socher. 2016.
\newblock Pointer sentinel mixture models.
\newblock \emph{arXiv preprint arXiv:1609.07843}.

\bibitem[{Mihaylov et~al.(2018)Mihaylov, Clark, Khot, and Sabharwal}]{mihaylov2018openbookqa}
Todor Mihaylov, Peter Clark, Tushar Khot, and Ashish Sabharwal. 2018.
\newblock \href {https://doi.org/10.18653/v1/D18-1260} {Can a suit of armor conduct electricity? a new dataset for open book question answering}.
\newblock In \emph{Proceedings of the 2018 Conference on Empirical Methods in Natural Language Processing}, pages 2381--2391, Brussels, Belgium. Association for Computational Linguistics.

\bibitem[{Nrusimha et~al.(2024)Nrusimha, Mishra, Wang, Alistarh, Panda, and Kim}]{nrusimha2024mitigating}
Aniruddha Nrusimha, Mayank Mishra, Naigang Wang, Dan Alistarh, Rameswar Panda, and Yoon Kim. 2024.
\newblock Mitigating the impact of outlier channels for language model quantization with activation regularization.
\newblock \emph{arXiv preprint arXiv:2404.03605}.

\bibitem[{Qin et~al.(2023)Qin, Li, Sun, Sun, Shen, Han, Wei, Lv, Yuan, Luo et~al.}]{qin2023transnormerllm}
Zhen Qin, Dong Li, Weigao Sun, Weixuan Sun, Xuyang Shen, Xiaodong Han, Yunshen Wei, Baohong Lv, Fei Yuan, Xiao Luo, et~al. 2023.
\newblock Scaling transnormer to 175 billion parameters.
\newblock \emph{arXiv preprint arXiv:2307.14995}.

\bibitem[{Sakaguchi et~al.(2020)Sakaguchi, Le~Bras, Bhagavatula, and Choi}]{Sakaguchi2020winogrande}
Keisuke Sakaguchi, Ronan Le~Bras, Chandra Bhagavatula, and Yejin Choi. 2020.
\newblock \href {https://doi.org/10.1609/aaai.v34i05.6399} {Winogrande: An adversarial winograd schema challenge at scale}.
\newblock \emph{Proceedings of the AAAI Conference on Artificial Intelligence}, 34(05):8732--8740.

\bibitem[{Sap et~al.(2019)Sap, Rashkin, Chen, Le~Bras, and Choi}]{sap2019siqa}
Maarten Sap, Hannah Rashkin, Derek Chen, Ronan Le~Bras, and Yejin Choi. 2019.
\newblock \href {https://doi.org/10.18653/v1/D19-1454} {Social {IQ}a: Commonsense reasoning about social interactions}.
\newblock In \emph{Proceedings of the 2019 Conference on Empirical Methods in Natural Language Processing and the 9th International Joint Conference on Natural Language Processing (EMNLP-IJCNLP)}, pages 4463--4473, Hong Kong, China. Association for Computational Linguistics.

\bibitem[{Schulz(1933)}]{schulz1933iterativenewtonschulz}
G{\"u}nther Schulz. 1933.
\newblock Iterative berechung der reziproken matrix.
\newblock \emph{ZAMM-Journal of Applied Mathematics and Mechanics/Zeitschrift f{\"u}r Angewandte Mathematik und Mechanik}, 13(1):57--59.

\bibitem[{Son et~al.(2024)Son, Park, Han, Kim, and Lee}]{son-etal-2024-prefixing-attn-sinks}
Seungwoo Son, Wonpyo Park, Woohyun Han, Kyuyeun Kim, and Jaeho Lee. 2024.
\newblock \href {https://doi.org/10.18653/v1/2024.emnlp-main.134} {Prefixing attention sinks can mitigate activation outliers for large language model quantization}.
\newblock In \emph{Proceedings of the 2024 Conference on Empirical Methods in Natural Language Processing}, pages 2242--2252, Miami, Florida, USA. Association for Computational Linguistics.

\bibitem[{Sun et~al.(2024)Sun, Chen, Kolter, and Liu}]{sun2024massiveactivations}
Mingjie Sun, Xinlei Chen, J~Zico Kolter, and Zhuang Liu. 2024.
\newblock Massive activations in large language models.
\newblock \emph{arXiv preprint arXiv:2402.17762}.

\bibitem[{Talmor et~al.(2019)Talmor, Herzig, Lourie, and Berant}]{talmor2019csqa}
Alon Talmor, Jonathan Herzig, Nicholas Lourie, and Jonathan Berant. 2019.
\newblock \href {https://doi.org/10.18653/v1/N19-1421} {{C}ommonsense{QA}: A question answering challenge targeting commonsense knowledge}.
\newblock In \emph{Proceedings of the 2019 Conference of the North {A}merican Chapter of the Association for Computational Linguistics: Human Language Technologies, Volume 1 (Long and Short Papers)}, pages 4149--4158, Minneapolis, Minnesota. Association for Computational Linguistics.

\bibitem[{Touvron et~al.(2023)Touvron, Lavril, Izacard, Martinet, Lachaux, Lacroix, Rozi{\`e}re, Goyal, Hambro, Azhar et~al.}]{touvron2023llama}
Hugo Touvron, Thibaut Lavril, Gautier Izacard, Xavier Martinet, Marie-Anne Lachaux, Timoth{\'e}e Lacroix, Baptiste Rozi{\`e}re, Naman Goyal, Eric Hambro, Faisal Azhar, et~al. 2023.
\newblock Llama: Open and efficient foundation language models.
\newblock \emph{arXiv preprint arXiv:2302.13971}.

\bibitem[{Tseng et~al.(2024)Tseng, Chee, Sun, Kuleshov, and De~Sa}]{tseng2024quipsharp}
Albert Tseng, Jerry Chee, Qingyao Sun, Volodymyr Kuleshov, and Christopher De~Sa. 2024.
\newblock \href {https://proceedings.mlr.press/v235/tseng24a.html} {{Q}u{IP}$\#$: Even better {LLM} quantization with hadamard incoherence and lattice codebooks}.
\newblock In \emph{Proceedings of the 41st International Conference on Machine Learning}, volume 235 of \emph{Proceedings of Machine Learning Research}, pages 48630--48656. PMLR.

\bibitem[{Vyas et~al.(2024)Vyas, Morwani, Zhao, Shapira, Brandfonbrener, Janson, and Kakade}]{vyas2024soap}
Nikhil Vyas, Depen Morwani, Rosie Zhao, Itai Shapira, David Brandfonbrener, Lucas Janson, and Sham Kakade. 2024.
\newblock Soap: Improving and stabilizing shampoo using adam.
\newblock \emph{arXiv preprint arXiv:2409.11321}.

\bibitem[{Wei et~al.(2022)Wei, Zhang, Zhang, Gong, Zhang, Zhang, Yu, and Liu}]{wei2022outliersuppression}
Xiuying Wei, Yunchen Zhang, Xiangguo Zhang, Ruihao Gong, Shanghang Zhang, Qi~Zhang, Fengwei Yu, and Xianglong Liu. 2022.
\newblock \href {https://proceedings.neurips.cc/paper_files/paper/2022/file/6f6db140de9c9f111b12ef8a216320a9-Paper-Conference.pdf} {Outlier suppression: Pushing the limit of low-bit transformer language models}.
\newblock In \emph{Advances in Neural Information Processing Systems}, volume~35, pages 17402--17414. Curran Associates, Inc.

\bibitem[{Wen et~al.(2024)Wen, Li, Wang, Hall, Liang, and Ma}]{wen2024warmupstabledecay}
Kaiyue Wen, Zhiyuan Li, Jason Wang, David Hall, Percy Liang, and Tengyu Ma. 2024.
\newblock Understanding warmup-stable-decay learning rates: A river valley loss landscape perspective.
\newblock \emph{arXiv preprint arXiv:2410.05192}.

\bibitem[{Xiao et~al.(2023)Xiao, Lin, Seznec, Wu, Demouth, and Han}]{xiao2023smoothquant}
Guangxuan Xiao, Ji~Lin, Mickael Seznec, Hao Wu, Julien Demouth, and Song Han. 2023.
\newblock \href {https://proceedings.mlr.press/v202/xiao23c.html} {{S}mooth{Q}uant: Accurate and efficient post-training quantization for large language models}.
\newblock In \emph{Proceedings of the 40th International Conference on Machine Learning}, volume 202 of \emph{Proceedings of Machine Learning Research}, pages 38087--38099. PMLR.

\bibitem[{Xiao et~al.(2024)Xiao, Tian, Chen, Han, and Lewis}]{xiao2024efficientstreaming}
Guangxuan Xiao, Yuandong Tian, Beidi Chen, Song Han, and Mike Lewis. 2024.
\newblock \href {https://openreview.net/forum?id=NG7sS51zVF} {Efficient streaming language models with attention sinks}.
\newblock In \emph{The Twelfth International Conference on Learning Representations}.

\bibitem[{Xu et~al.(2020)Xu, Lee, Chen, Choi, Hechtman, and Wang}]{xu2020fsdp}
Yuanzhong Xu, HyoukJoong Lee, Dehao Chen, Hongjun Choi, Blake Hechtman, and Shibo Wang. 2020.
\newblock Automatic cross-replica sharding of weight update in data-parallel training.
\newblock \emph{arXiv preprint arXiv:2004.13336}.

\bibitem[{Yang et~al.(2024{\natexlab{a}})Yang, Yang, Hui, Zheng, Yu, Zhou, Li, Li, Liu, Huang, Dong, Wei, Lin, Tang, Wang, Yang, Tu, Zhang, Ma, Xu, Zhou, Bai, He, Lin, Dang, Lu, Chen, Yang, Li, Xue, Ni, Zhang, Wang, Peng, Men, Gao, Lin, Wang, Bai, Tan, Zhu, Li, Liu, Ge, Deng, Zhou, Ren, Zhang, Wei, Ren, Fan, Yao, Zhang, Wan, Chu, Liu, Cui, Zhang, and Fan}]{qwen2}
An~Yang, Baosong Yang, Binyuan Hui, Bo~Zheng, Bowen Yu, Chang Zhou, Chengpeng Li, Chengyuan Li, Dayiheng Liu, Fei Huang, Guanting Dong, Haoran Wei, Huan Lin, Jialong Tang, Jialin Wang, Jian Yang, Jianhong Tu, Jianwei Zhang, Jianxin Ma, Jin Xu, Jingren Zhou, Jinze Bai, Jinzheng He, Junyang Lin, Kai Dang, Keming Lu, Keqin Chen, Kexin Yang, Mei Li, Mingfeng Xue, Na~Ni, Pei Zhang, Peng Wang, Ru~Peng, Rui Men, Ruize Gao, Runji Lin, Shijie Wang, Shuai Bai, Sinan Tan, Tianhang Zhu, Tianhao Li, Tianyu Liu, Wenbin Ge, Xiaodong Deng, Xiaohuan Zhou, Xingzhang Ren, Xinyu Zhang, Xipin Wei, Xuancheng Ren, Yang Fan, Yang Yao, Yichang Zhang, Yu~Wan, Yunfei Chu, Yuqiong Liu, Zeyu Cui, Zhenru Zhang, and Zhihao Fan. 2024{\natexlab{a}}.
\newblock Qwen2 technical report.
\newblock \emph{arXiv preprint arXiv:2407.10671}.

\bibitem[{Yang et~al.(2024{\natexlab{b}})Yang, Yang, Zhang, Hui, Zheng, Yu, Li, Liu, Huang, Wei, Lin, Yang, Tu, Zhang, Yang, Yang, Zhou, Lin, Dang, Lu, Bao, Yang, Yu, Li, Xue, Zhang, Zhu, Men, Lin, Li, Xia, Ren, Ren, Fan, Su, Zhang, Wan, Liu, Cui, Zhang, and Qiu}]{qwen2.5}
An~Yang, Baosong Yang, Beichen Zhang, Binyuan Hui, Bo~Zheng, Bowen Yu, Chengyuan Li, Dayiheng Liu, Fei Huang, Haoran Wei, Huan Lin, Jian Yang, Jianhong Tu, Jianwei Zhang, Jianxin Yang, Jiaxi Yang, Jingren Zhou, Junyang Lin, Kai Dang, Keming Lu, Keqin Bao, Kexin Yang, Le~Yu, Mei Li, Mingfeng Xue, Pei Zhang, Qin Zhu, Rui Men, Runji Lin, Tianhao Li, Tingyu Xia, Xingzhang Ren, Xuancheng Ren, Yang Fan, Yang Su, Yichang Zhang, Yu~Wan, Yuqiong Liu, Zeyu Cui, Zhenru Zhang, and Zihan Qiu. 2024{\natexlab{b}}.
\newblock Qwen2.5 technical report.
\newblock \emph{arXiv preprint arXiv:2412.15115}.

\bibitem[{Zellers et~al.(2019)Zellers, Holtzman, Bisk, Farhadi, and Choi}]{zellers2019hellaswag}
Rowan Zellers, Ari Holtzman, Yonatan Bisk, Ali Farhadi, and Yejin Choi. 2019.
\newblock \href {https://doi.org/10.18653/v1/P19-1472} {{H}ella{S}wag: Can a machine really finish your sentence?}
\newblock In \emph{Proceedings of the 57th Annual Meeting of the Association for Computational Linguistics}, pages 4791--4800, Florence, Italy. Association for Computational Linguistics.

\bibitem[{Zhang et~al.(2024)Zhang, Zeng, Wang, and Lu}]{zhang2024tinyllama}
Peiyuan Zhang, Guangtao Zeng, Tianduo Wang, and Wei Lu. 2024.
\newblock \href {https://arxiv.org/abs/2401.02385} {Tinyllama: An open-source small language model}.
\newblock \emph{Preprint}, arXiv:2401.02385.

\bibitem[{Zhang et~al.(2022)Zhang, Roller, Goyal, Artetxe, Chen, Chen, Dewan, Diab, Li, Lin et~al.}]{zhang2022opt}
Susan Zhang, Stephen Roller, Naman Goyal, Mikel Artetxe, Moya Chen, Shuohui Chen, Christopher Dewan, Mona Diab, Xian Li, Xi~Victoria Lin, et~al. 2022.
\newblock Opt: Open pre-trained transformer language models.
\newblock \emph{arXiv preprint arXiv:2205.01068}.

\end{thebibliography}

\newpage
\appendix
\section{Appendix}

\subsection{Additional Training Details}

For training data, we adopt a corpus composition similar to \citet{allal2025smollm2}, leveraging carefully validated mixture of high-quality datasets. This corpus comprises FineWeb-Edu~\citep{lozhkov2024fineweb-edu}, FineMath~\citep{allal2025smollm2}, Cosmopedia~\citep{benallal2024cosmopedia}, and Python codes sampled from the StarCoder~\citep{li2023starcoder} training set. The selection of this data mixture enables direct comparison with existing benchmarks while ensuring robust evaluation across various downstream tasks.

Our training infrastructure utilizes a TPU v4-512 Pod Slice, enabling efficient distributed training at scale. We conduct comparative experiments between the standard Adam optimizer with a learning rate of $5\times10^{-3}$ and the Muon optimizer with a learning rate of $5\times10^{-4}$. The training configuration maintains a batch size of 4 million tokens with a sequence length of 2048 tokens, applying weight decay of 0.01 across all experiments. We adopt trapezoidal learning rate scheduling \citep{hagele2024trapezoidal, wen2024warmupstabledecay}, wherein the learning rate increases linearly from zero to its maximum value over the first 5 billion tokens, maintains this peak throughout the majority of training, and subsequently decays to zero during the final 20\% steps.

To achieve optimal training throughput, we implement Fully-Sharded Data-Parallel (FSDP)~\citep{xu2020fsdp} with parameters distributed across 16 accelerator cores. For Muon optimization, we develop a distributed variant that partitions gradients across 8 dedicated optimizer-parallel ranks, where Newton-Schulz iterations are performed independently on each rank. This parallelization strategy enables efficient orthogonalization of large gradient matrices without communication bottlenecks.

\subsection{Comprehensive Benchmark Results for Open-Source LLMs}
\begin{table*}[t]
\centering
\resizebox{1.0\textwidth}{!}{%
\begin{tabular}{l|cc|cccccccccc|c}
\toprule
\textbf{Model} & \textbf{Params.} & \textbf{Tokens} & \textbf{ARC} & \textbf{CSQA} & \textbf{GSM8K} & \textbf{HS} & \textbf{MMLU} & \textbf{OBQA} & \textbf{PIQA} & \textbf{SIQA} & \textbf{TQA} & \textbf{WG} & \textbf{Avg.} \\

\midrule

Pythia\nocite{pythia} & 1.4B & 0.3T & 41.3 & 35.4 & 2.4 & 50.8 & 31.3 & 34.6 & 71.1 & 43.5 & 9.2 & 55.2 & 37.5 \\

TinyLlama\nocite{zhang2024tinyllama} & 1.1B & 2T & 36.5 & 25.4 & 1.7 & 54.0 & 32.6 & 23.0 & 70.3 & 41.3 & 23.5 & 50.0 & 35.8 \\

OPT\nocite{zhang2022opt} & 1.3B & 0.3T & 39.3 & 40.0 & 0.9 & 52.2 & 29.6 & 35.8 & 71.0 & 42.3 & 11.1 & 53.3 & 37.6 \\

OLMo\nocite{groeneveld2024olmo} & 1.2B & 3T & 44.2 & 40.4 & 1.7 & 60.4 & 31.9 & 37.8 & 75.2 & 44.1 & 17.6 & 53.4 & 40.7 \\

Mobile\textsc{LLaMA}\nocite{chu2023mobilellama} & 1.4B & 1.3T & 42.7 & 37.0 & 2.0 & 54.2 & 31.8 & 34.4 & 73.3 & 43.0 & 24.5 & 55.4 & 39.8 \\

Qwen 1.5\nocite{qwen1.5} & 1.8B & 2.4T & 46.9 & 32.9 & 34.2 & 59.5 & 33.1 & 37.2 & 74.3 & 44.5 & 18.8 & 57.9 & 43.9 \\

Qwen 2\nocite{qwen2} & 1.5B & 7T & 48.2 & 31.0 & 58.1 & 63.9 & 37.4 & 36.8 & 75.4 & 44.2 & 24.0 & 59.2 & 47.8 \\

Qwen 2.5\nocite{qwen2.5} & 1.5B & -- &  58.8 & 34.3 & \textbf{61.6} & 66.5 & 40.3 & 39.6 & 75.7 & \textbf{44.9} & 20.6 & 59.4 & \textbf{50.2} \\

\textsc{LLaMA} 3.2\nocite{dubey2024llama} & 1.2B & -- & 49.2 & 41.1 & 6.0 & 61.3 & 36.3 & 39.0 & 74.9 & 43.5 & 20.7 & 58.1 & 43.0 \\

Stable LM 2\nocite{bellagente2024stablelm2} & 1.6B & 2T & 53.5 & 34.6 & 19.3 & 66.7 & 36.0 & 37.0 & 76.8 & 43.5 & \textbf{35.6} & 59.2 & 46.2 \\

SmolLM\nocite{allal2024SmolLM} & 1.7B & 1T & 59.7 & 38.0 & 6.8 & 63.0 & 39.4 & \textbf{42.8} & 76.0 & 44.1 & 25.8 & 54.6 & 45.0 \\

SmolLM 2\nocite{allal2025smollm2} & 1.7B & 11T & \textbf{60.4} & \textbf{43.6} & 32.6 & \textbf{68.7} & \textbf{41.3} & 42.4 & \textbf{77.6} & 43.4 & 27.1 & \textbf{60.1} & 49.7 \\

\midrule
\multicolumn{14}{c}{\textbf{From Scratch}} \\
\midrule

Adam & 1.4B & 1T & 59.5 & 40.6 & 14.5 & 64.0 & 39.5 & 41.0 & 76.1 & 43.6 & 23.9 & 56.6 & 45.9 \\

Muon (\textbf{OSP}) & 1.4B & 1T & 57.5 & 37.6 & 10.5 & 61.3 & 38.5 & 40.4 & 75.5 & 44.4 & 22.4 & 55.8 & 44.4 \\
\bottomrule

\end{tabular}
}
\caption{
Performance evaluation of 12 open-source large language models and our two models trained from scratch, assessed without quantization across 10 benchmark tasks. Model parameters (\textbf{Params.}) represent total trainable parameters, while \textbf{Tokens} indicate training dataset size. Evaluation benchmarks include CommonsenseQA (CSQA), HellaSwag (HS), OpenBookQA (OBQA), TriviaQA (TQA), and WinoGrande (WG), among others. Results provide comprehensive baseline performance metrics across diverse reasoning and knowledge tasks, demonstrating the comparative performance of our trained models against established baselines.
}
\label{tab:result-on-ollms-w16a16}
\end{table*}
\begin{figure*}[t]
\centering
\hspace*{-1em}
\includegraphics[width=0.95\textwidth]{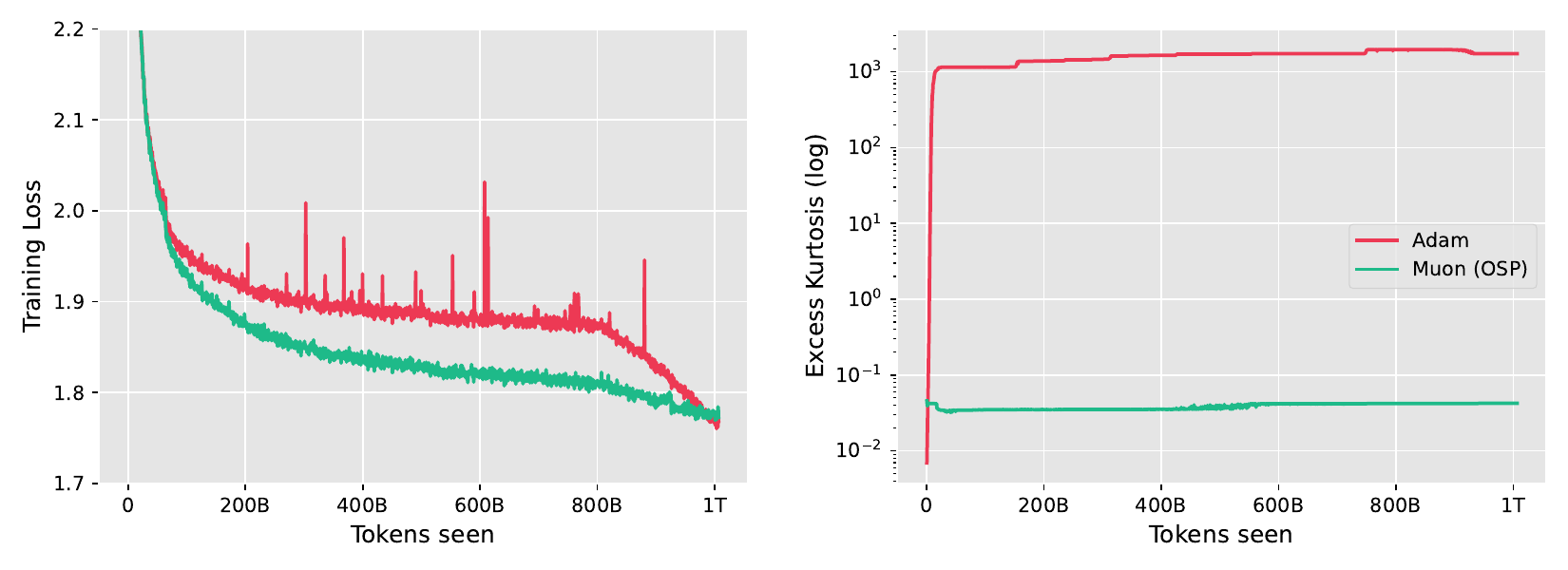}
\vspace*{-1em}
\caption{
Training dynamics over 1 trillion tokens demonstrating production-scale viability of our framework. The loss (left) and excess kurtosis evolution (right) compare Adam baseline against  complete {OSP} implementation. Results confirm that the {OSP} maintains consistently low kurtosis values throughout extended training, validating the framework's effectiveness at production scale while achieving competitive convergence characteristics.
}
\label{fig:1T-training-log}
\end{figure*}

Table~\ref{tab:result-on-ollms-w16a16} presents the performance across 10 benchmarks without quantization. In particular, the model trained under our framework achieves comparable performance to the open-source models trained with Adam optimizer, confirming the successful application to trillion-token scale training.

\subsection{Training Dynamics Over 1T Token Scale}

Figure~\ref{fig:1T-training-log} illustrates the evolution of training loss and excess kurtosis throughout the one-trillion token training process, mirroring our ablation study methodology. To ensure fair comparison, we trained a standard Adam-based model under identical conditions.

\subsection{Detailed Weight and Activation Distributions}

\begin{figure*}[t]
\centering
\includegraphics[width=0.8\linewidth]{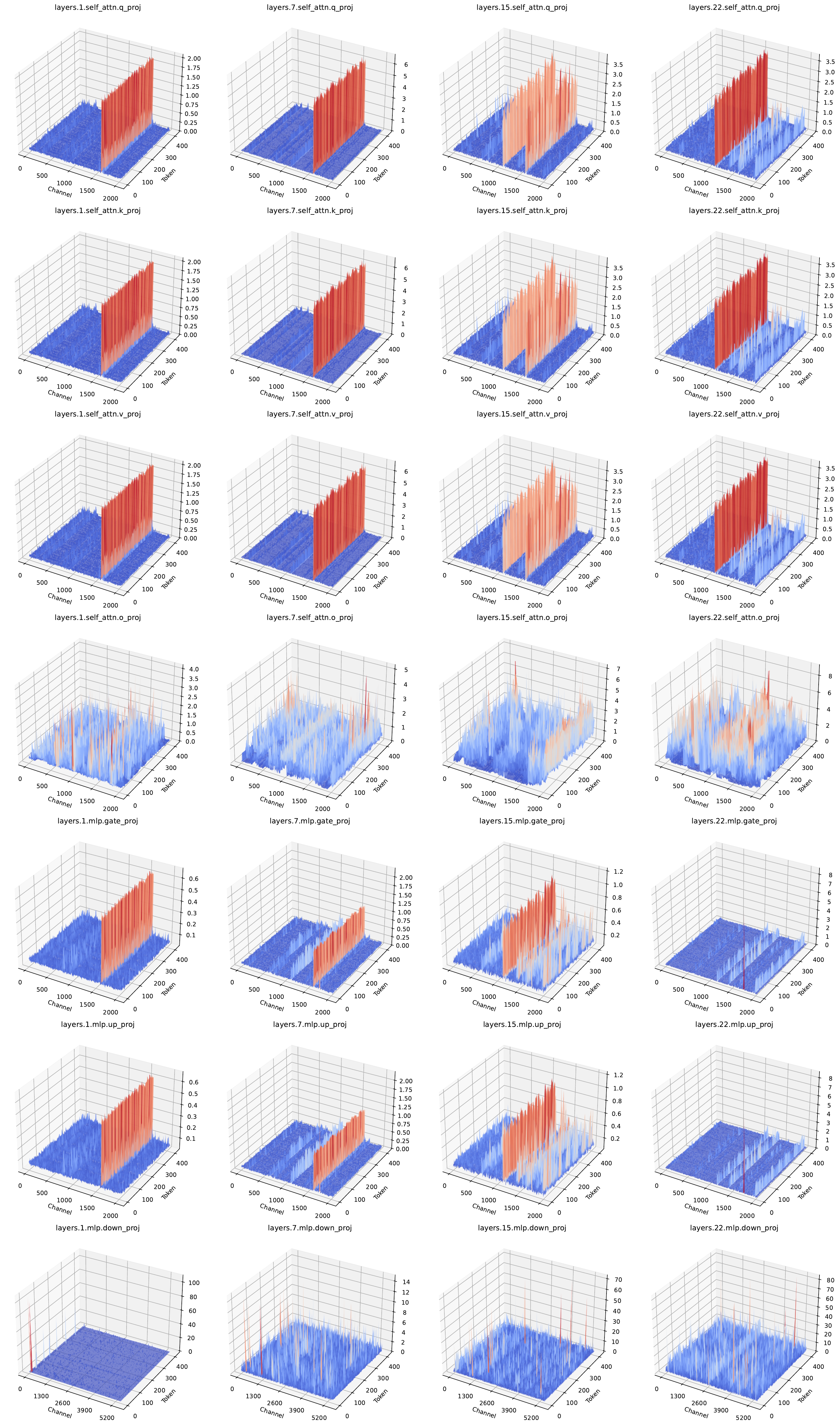}
\caption{
Activation distribution visualization of models trained with Adam optimizer across 1 trillion training tokens. The histograms display input activation distributions to Multi-Head Self-Attention (MHSA) and Feed-Forward Network (FFN) layers at four transformer block depths: 1st, 7th, 15th, and 22nd layers. Results demonstrate the evolution of activation patterns across network depth and illustrate the characteristic input distribution behavior produced by standard Adam optimization in both attention and feed-forward layers.
}
\label{fig:1T-act-adam-hist}
\end{figure*}

\begin{figure*}[t]
\centering
\includegraphics[width=0.8\linewidth]{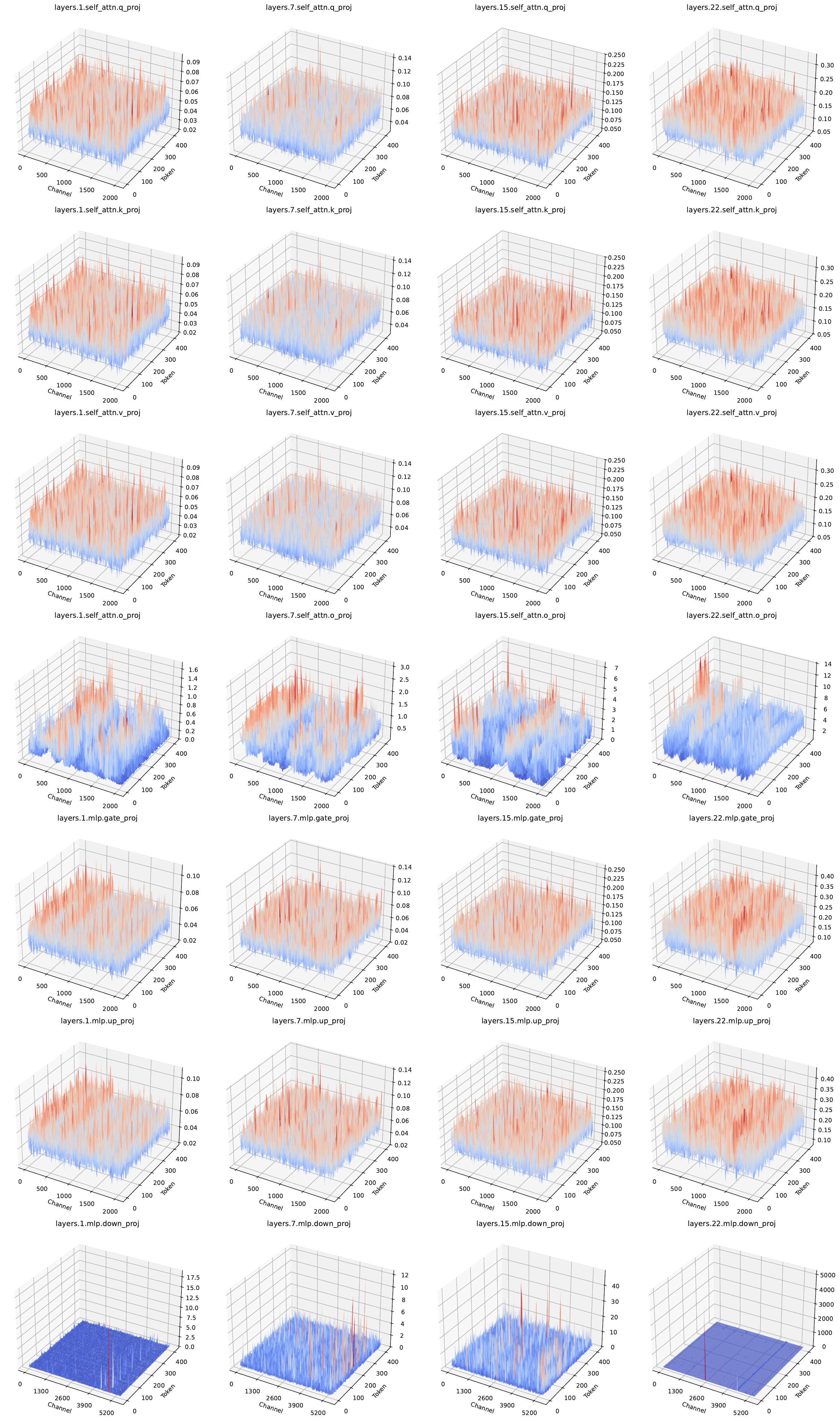}
\caption{
Activation distribution visualization of models trained with \textbf{OSP} across 1 trillion training tokens. The histograms display input activation distributions to Multi-Head Self-Attention (MHSA) and Feed-Forward Network (FFN) layers at four transformer block depths: 1st, 7th, 15th, and 22nd layers. Results demonstrate the evolution of activation patterns across network depth and illustrate the distinctive input distribution characteristics achieved through the \textbf{OSP} framework in both attention and feed-forward layers.
}
\label{fig:1T-act-muon-hist}
\end{figure*}

\begin{figure*}[t]
\centering
\includegraphics[width=0.8\linewidth]{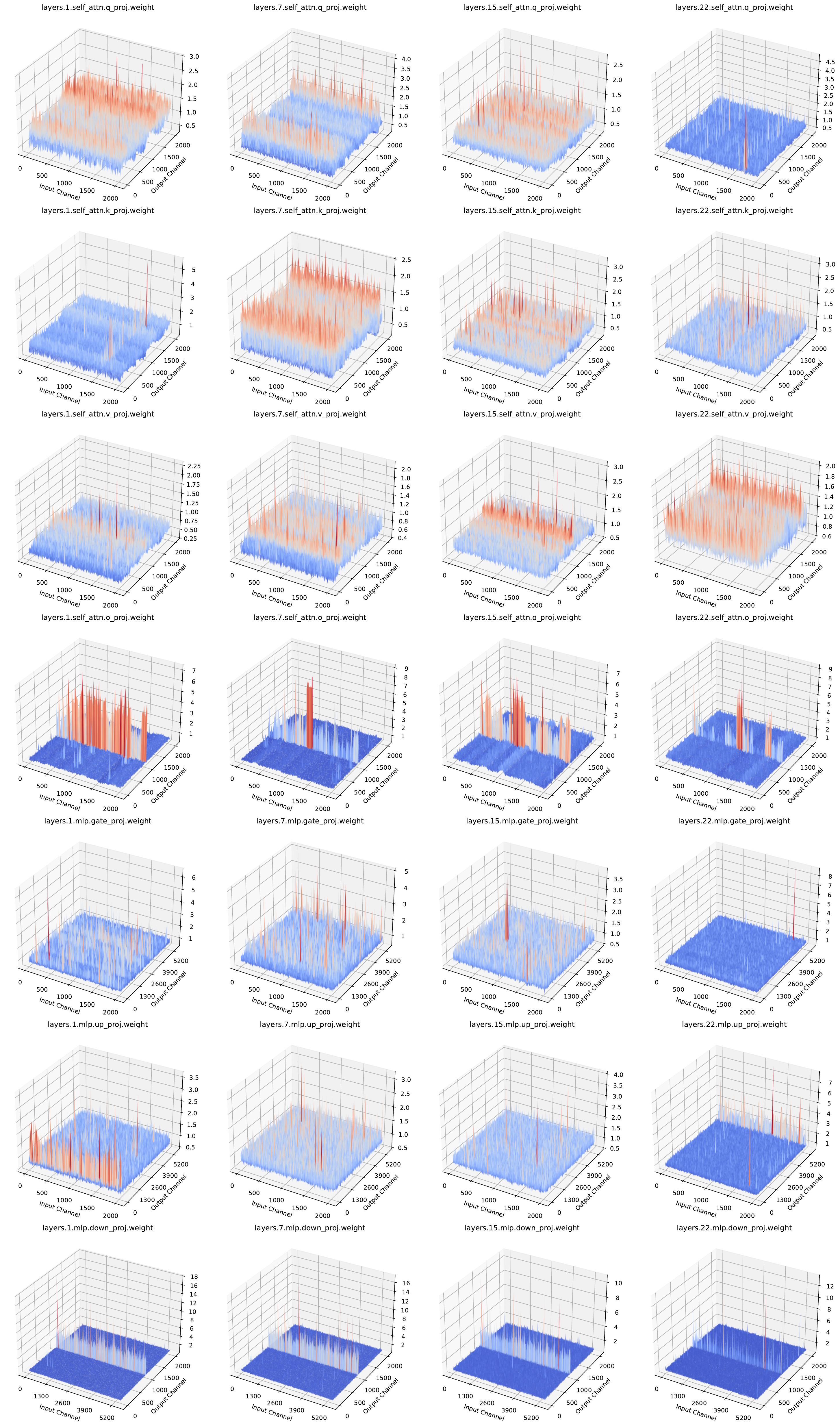}
\caption{
Weight distribution visualization of models trained with Adam optimizer across 1 trillion training tokens. The histograms display weight distributions within Multi-Head Self-Attention (MHSA) and Feed-Forward Network (FFN) layers at four transformer block depths: 1st, 7th, 15th, and 22nd layers. Results illustrate the evolution of weight distributions across network depth and demonstrate the characteristic patterns produced by standard Adam optimization in both attention and feed-forward layers.
}
\label{fig:1T-weight-adam-hist}
\end{figure*}

\begin{figure*}[t]
\centering
\includegraphics[width=0.8\linewidth]{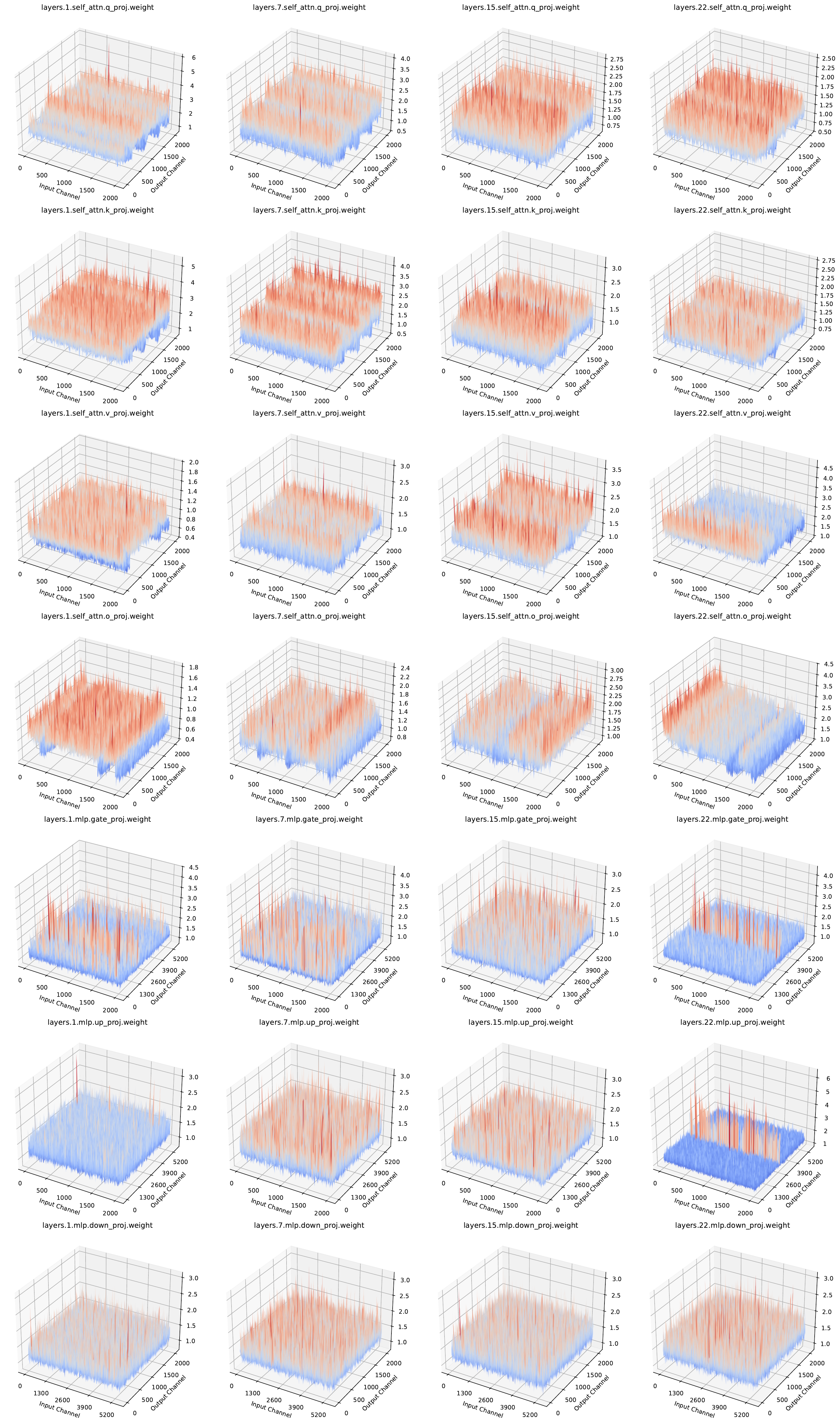}
\caption{
Weight distribution visualization of model trained with \textbf{OSP} across 1 trillion training tokens. The histograms display weight distributions within Multi-Head Self-Attention (MHSA) and Feed-Forward Network (FFN) layers at four transformer block depths: 1st, 7th, 15th, and 22nd layers. Results illustrate the evolution of weight distributions across network depth and demonstrate the characteristic patterns induced by the \textbf{OSP} framework in both attention and feed-forward layers.
}
\label{fig:1T-weight-muon-hist}
\end{figure*}

For a comprehensive view of activation and weight distributions, we provide detailed histograms in Figures~\ref{fig:1T-act-adam-hist}, \ref{fig:1T-act-muon-hist}, \ref{fig:1T-weight-adam-hist}, and \ref{fig:1T-weight-muon-hist}.

\end{document}